\documentclass[preprint,12pt]{elsarticle}

\usepackage{amssymb}
\usepackage{amsmath}
\usepackage{subcaption}
\usepackage{url}
\usepackage{makecell}
\usepackage{float}
\usepackage{xcolor}

\journal{European Journal of Operations Research}

\begin{document}

\begin{frontmatter}



\title{The more the merrier:\\logical and multistage processors in credit scoring}


\author[inst3]{Arturo P\'erez-Peralta}

\author[inst2,inst3]{Sandra Ben\'itez-Peña}
\author[inst2,inst3]{Rosa E. Lillo }

\affiliation[inst2]{organization={Department of Statistics. Universidad Carlos III de Madrid},
            country={Spain}}
\affiliation[inst3]{organization={uc3m-Santander Big Data Institute. Universidad Carlos III de Madrid},
            country={Spain}}

\begin{abstract}
Machine Learning (ML) algorithms are ubiquitous in key decision-making contexts such as organizational justice or healthcare, which has spawned a great demand for fairness in these procedures. In this paper we focus on the application of fair ML in finance, more concretely on the use of fairness techniques on credit scoring. This paper makes two contributions. On the one hand, it addresses the existent gap concerning the application of established methods in the literature to the case of multiple sensitive variables through the use of a new technique called \emph{logical processors} (LP). On the other hand, it also explores the novel method of \emph{multistage processors} (MP) to investigate whether the combination of fairness methods can work synergistically to produce solutions with improved fairness or accuracy. Furthermore, we examine the intersection of these two lines of research by exploring the integration of fairness methods in the multivariate case. The results are very promising and suggest that logical processors are an appropriate way of handling multiple sensitive variables. Furthermore, multistage processors are capable of improving the performance of existing methods.
\end{abstract}


\end{frontmatter}


\section{Introduction}
\label{sec:introduction}

In the last decades, institutions have been increasingly relying on artificial intelligence (AI) and machine learning (ML) to aid in decision-making. From personnel selection \cite{kuncel2014hiring} to health care \cite{park2020artificial}, intelligent systems have become ubiquitous in critical contexts \cite{cite:surveyapplications}, which is why there is a growing demand for socially aware models and the reason of being of fair AI  \cite{cutler2019everyday}. Furthermore, the interplay between discrimination and calibration suggests that building a model avoiding spurious relationships between variables may increase reliability \cite{diamond1992price}. This paper will focus on the application of fair ML models in a financial context to address the problem of credit scoring, which plays a key role in loan approval \cite{mester1997s}.\\
Although a plethora of metrics and models have been proposed in the literature for bias mitigation, there are still many open challenges surrounding this topic. More concretely, this work is interested in exploring two particular research gaps. On the one hand, there is a demand for methods that handle multiple sensitive variables both from ethical and legal frameworks \cite{caton2024fairness}. Furthermore, there are concerns about the unique discrimination that some individuals experience due to their belonging to the intersection of protected groups \cite{malloy2010standards}. On the other hand, existing fairness processors act during a single stage of the ML pipeline: they are implemented either while treating the data (pre-processors), during training (in-processors) or after predicting (post-processors) \cite{pagano2023bias}. To our knowledge, there is no literature on hybrid methods involving multiple stages of the algorithm-making process. This may be a fruitful endeavor in order to better understand the fairness-performance trade-offs \cite{cite:multiplemotivation}.\\ 
In order to attack these questions this paper is structured as follows. First, in Section~\ref{sec:background}, we will review the theoretical background necessary to understand the ideas we present, explaining the metrics that will be used to quantify bias and the processors that have been developed to address discrimination \cite{cite:FAIRMLBOOK}. Then, in Section~\ref{sec:extending}, the two main ideas behind this paper will be introduced. On the one hand, we propose \emph{logical processors}, a method based on the use of bitwise operations to reduce the problem of multiple sensitive variables to the univariate case. This allows greater flexibility when choosing and designing fairness processors and adresses one of the biggest research gaps in the current literature; namely, the generalization of fairness methods to the multivariate case \cite{cite:POSTOPTREJECT}. On the other hand, we will explore \emph{multistage processors}, that is, the use of multiple fairness techniques that affect different parts of the ML pipeline in an attempt to find synergistic combinations that may lead to improved fairness or accuracy. This notion of hybrid methods has already been proposed \cite{cite:multiplemotivation}, but to our knowledge no results have been produced in this direction. Furthermore, the interplay of logical and multistage processors is analyzed in an attempt to achieve better results from the perspective of fairness. Section~\ref{sec:setup}  compiles an extensive empirical study to understand the feasibility and efficiency of the ideas presented in Section 3. The analysis will be divided into two parts. First,  a simulation study similar to the one proposed in \cite{cite:INADVERSARIAL} to check how our methods fare in a controlled environment. Then, a case study using the German\footnote{\url{https://archive.ics.uci.edu/dataset/144/statlog+german+credit+data}} data set to get a better  grasp of the trade-offs involved in the use of the new techniques. Finally, Section~\ref{sec:further_research} presents the conclusions and outlines directions for future research.

\section{The Foundations of Fairness }
\label{sec:background}

This Section lays down the theoretical foundation of the paper following \cite{cite:FAIRMLBOOK}. The review will formulate the problem of ML fairness in the broader context of statistical classification and visit the metrics that have been proposed in the literature to tackle the problem of measuring bias and prejudice.

\subsection{The Classification problem under the fair pointwise}
\label{subsec:class}

Suppose a data set $D = (X, Y, A)$ comprised of a set of attributes, $X$, a target variable, $Y$, and a set of sensitive variables, $A$. The work \cite{cite:FAIRMLBOOK} assumes that both the response and the sensitive variables are binary; that is, $A, Y \in \{0, 1\}$. Furthermore, we restrict ourselves to the case where $A$ represents a single sensitive attribute. The goal of statistical classification is then to develop a model, $\hat{Y} : dom(X) \longrightarrow \{0,1 \}$, which receives the name of classifier, while maximizing its accuracy (or, equivalently, while minimizing its error rate),
\begin{equation}
    \text{max}\ \mathbb{P}(\hat{Y} = Y ) \equiv \text{min} \ \mathbb{P} (\hat{Y} \neq Y ),
\end{equation}
where $\mathbb{P}$ denotes the probability of an event. When refering to empirical probabilities the $\widehat{\mathbb{P}}$ symbol will be used. However, in the context of fairness it makes more sense to maximize the balanced accuracy, that is, the mean group accuracy,
\begin{equation}
\textit{balanced accuracy} = \frac{1}{|A|} \sum_{a\in A}  \mathbb{P}(\hat{Y} = Y | A = a).
\label{balacc}
\end{equation}
Other important performance metrics are the entries of the confusion matrix:
\begin{itemize}
    \item \textbf{True positive ratio:} $TPR = \mathbb{P}(\hat{Y}  = 1 \mid Y = 1)$.
    \item \textbf{True negative ratio:} $TNR = \mathbb{P}(\hat{Y}  = 0 \mid Y = 0)$.
    \item \textbf{False positive ratio:} $FPR = \mathbb{P}(\hat{Y}  = 1 \mid Y = 0)$.
    \item \textbf{False negative ratio:} $FNR = \mathbb{P}(\hat{Y}  = 0 \mid Y = 1)$.
\end{itemize}
In the context of credit scoring the response variable $Y$ represents whether an individual will repay a loan. When the individual is predicted to default they are said to have bad credit score or bad credit risk, and when they are predicted to repay the loan they are said to have good credit score or good credit risk. The broader interpretation of the relevant variables along with their values can be found in Table \ref{tab:interpretations}. In any case, the model $\hat{Y}$ is obtained by thresholding a score $R: dom(X) \longrightarrow \mathbb{R}$ that ranks the different clients according to their probability of default. Formally, a threshold $\tau$ is set and the classifier is built by thresholding the score $R$, that is, $\hat{Y} = \mathbf{1}$ if $R(X) > \tau$ and $\hat{Y} = \mathbf{0}$ otherwise.\\

\begin{table}[h!]
    \centering
    \begin{tabular}{c|c}
        Expression & Interpretation  \\
        \hline
         $A = 0$ & The individual belongs to the privileged group  \\
         \hline
         $A = 1$ & The individual belongs to the unprivileged group  \\
         \hline
         $Y = 0$ & The individual defaults (bad outcome)  \\
         \hline
         $Y = 1$ & The individual repays the loan (good outcome) \\
        \hline
         $\hat{Y} = 0$ & The individual is predicted to default (bad credit risk)\\
         \hline
         $\hat{Y} = 1$ & The individual is predicted to repay the loan (good credit risk) \\
    \end{tabular}
    \caption{Interpretation of the different values of the sensitive attribute and the response}
    \label{tab:interpretations}
\end{table}

Now the only question that remains is how to measure bias and prejudice.

\subsection{Fairness criteria}
\label{subsec:metrics}
Many fairness metrics have been proposed in the literature. However, most of them can be reduced to just three \cite{cite:FAIRMLBOOK}: Independence, separation and sufficiency.

\subsubsection{Independence}

A score $R(X)$ is said to satisfy independence at a threshold $\tau$ if

\begin{equation}
\mathbb{P}[R(X) > \tau \mid A = 1 ] = \mathbb{P}[R(X) > \tau \mid A = 0 ].
\label{INDEPENDENCE}
\end{equation}
That is, independence requires that the distribution of the classifier is independent of $A$. However, this assumption might be counterproductive if the underlying distribution of the data is not independent itself, in which case enforcing this metric may lead to an inaccurate model \cite{cite:EJORSurvey}. In order to avoid discrimination while retaining this metric, some institutions propose to bound the quotient of the quantities in \eqref{INDEPENDENCE}, generally by $0.8$ which is commonly referred to as the \emph{80\% rule} \cite{rubin1978uniform}. Nonetheless, the use of a fairness metric should be determined by the use case, although we will return to this question at the end of this Section.\\
Independence can be measured by the absolute value of the deviation in the above probabilities,

\begin{equation}
IND = | \mathbb{P}[R(X) > \tau \mid A = 1 ] - \mathbb{P}[R(X) > \tau \mid A = 0 ]|.
\label{IND}
\end{equation}

\noindent Therefore, a positive value for $IND$ implies a divergence from the equality in \eqref{INDEPENDENCE}. Hence, the closer $IND$ is to zero, the lower the discrimination is.

\subsubsection{Separation}
A score $R(X)$ is said to satisfy separation at a threshold $\tau$ if
\begin{subequations}
\begin{align}
    \mathbb{P}[R(X) > \tau \mid Y = 0, A = 1 ] = \mathbb{P}[R(X) > \tau \mid Y = 0, A = 0 ], \label{SEPARATION1} \\
   \mathbb{P}[R(X) \leq \tau \mid Y = 1, A = 1 ] = \mathbb{P}[R(X) \leq \tau \mid Y = 1, A = 0 ].
    \label{SEPARATION2}
\end{align}
\end{subequations}
That is, a score fulfils separation if all groups have equal error rates. The positive outcome is not assumed to be equally distributed, but ideally the error rate across population groups are equalized across the different values of the response; that is, a classifier that aims for separation does not try to improve global accuracy at the cost of misclassifying individuals from the minority group. This metric also receives the name \emph{equal odds}. A relaxation can be found in \emph{equal opportunity} in which only equation \eqref{SEPARATION1} is satisfied \cite{cite:POSTEQODDS}.\\ 
Separation can be measured with the unweighted average of the absolute value of the deviation of the false positive and negative rates,
\begin{equation}
SP = \frac{1}{2} | (FPR_{A = 1} - FPR_{A = 0} ) + (FNR_{A = 1} - FNR_{A = 0} )|.
\label{SP}
\end{equation}
$SP$ has a similar interpretation to $IND$: the close it is to zero, the closer the classifier is to achieve separation and the lower the discrimination.\\

\subsubsection{Sufficiency}
A score $R(X)$ is said to satisfy sufficiency at a threshold $\tau$ if
\begin{equation}
\mathbb{P}[Y = 1 \mid R(X) > \tau, A = 1 ] = \mathbb{P}[Y = 1 \mid R(X) > \tau, A = 0 ].
\label{SUFFICIENCY}
\end{equation}
That is, it requires that all the information of the target variable is contained in the score. 
In any case, sufficiency can be measured as the absolute value of the deviation of the above probabilities,
\begin{equation}
SF = |\mathbb{P}[Y = 1 \mid R(X) > \tau, A = 1 ] - \mathbb{P}[Y = 1 \mid R(X) > \tau, A = 0 ] |.
\label{SF}
\end{equation}

\subsubsection{What metric to use?}

The choice of a fairness metric depends on the use case. According to \cite{cite:EJORSurvey}, both independence and sufficiency are inappropriate in the context of credit scoring, and the use of separation is recommended.\\
Independence fails to account for the unequal distribution of loan payments. Moreover, the consequence of a default are more severe for the client than the institution, so enforcing independence would exacerbate discrimination rather than solving it. Sufficiency, on the other hand, allows discrimination through separation. Finally, separation addresses these shortcomings and takes into account the different misclassification costs, which makes it particularly appropriate for credit scoring.
    
\subsection{Fairness processors}
\label{sec:processors}

This Section will introduce the different fairness processors that have been considered for this study. These can be classified by their position in the ML pipeline. Broadly, there are three categories: pre-processors, in-processors and post-processors \cite{pagano2023bias}. The chosen methods will be briefly explained to provide the necessary context.

\subsubsection{Pre-Processing}
\label{subsec:pre}

Fair pre-processors transform the underlying data in order to mitigate any existing bias, the idea being that a classifier that is trained on unbiased data will not learn any discriminatory pattern.\\
The \emph{Reweighing} method described in \cite{cite:PREREWEIGHIN} aims to achieve independence by resampling the data with replacement using weights. These weights assume that the underlying data is independent, and they adjust the sampling frequencies so as to obtain a new data set that does not deviate from the previous assumption. We can estimate these weights using the empirical proportions 
\begin{equation}
 W(A = a \mid Y = y) = \frac{\mathbb{P}_{exp}(A = a, Y =y)}{\mathbb{P}_{act}(A = a, Y = y)} \approx \frac{\hat{\mathbb{P}}(A = a) \hat{\mathbb{P}}(Y =y)}{\hat{\mathbb{P}}(A = a, Y = y)},
\label{REWEIGHIN}
\end{equation}
where $\mathbb{P}_{exp}$ denotes the expected probability under the independence assumption, and $\mathbb{P}_{act}$ denotes the actual probability.

The second pre-processor that will be considered is the \emph{Disparate Impact Remover} \cite{cite:PREDI}. This procedure repairs a data set by transforming it into a new version, $\overline{D}$, which removes all information of the sensitive variable from the rest of the attributes. The repair criteria is based on the median distribution, $M$, of the $a$-conditional quantile functions of each attribute, $F_a^{-1}$,
\begin{equation} 
\overline{x} = F_M(F_a^{-1}(x)),\quad  \text{where } A(x) = a, \,  F_M^{-1} (u) = \text{median}\{ F_a^{-1}(u) \mid a \in A\}. 
\label{DISPARATEIMPACT}
\end{equation}
However, this procedure greatly compromises the predictive capabilities of any classifier that learns on this data. This can be remedied by the use a distribution generated as a geometric interpolation between the median quantile function, $F_M^{-1}$, and the corresponding $a$-conditional quantile, $F_a^{-1}$. This process is controled by a parameter $\lambda\in [0,1]$ that allows some control on the fairness-accuracy trade-off.

\subsubsection{In-Processing}
\label{subsec:in}

In-processors actively manipulate the training stage in order to produce a fair classifier. This is the idea behind the adversarial debiasing \cite{cite:INADVERSARIAL}. It is based on the introduction of an \emph{adversary} that will try to predict the sensitive attribute using some information. This information depends on the fairness metric of choice. The loss function of the original classifier, 
$L$, is modified to prevent the maximization of the loss of the adversary, $L_A$. The goal of this modification is to develop a predictor that maintains high accuracy while minimizing the amount of information it conveys about sensitive variables. The rule that it is used to update the weights is
\begin{equation}
\nabla _W L - \text{proj}_{\nabla_W L_A} \nabla_WL - \alpha \nabla_W L_A,
\label{ADVERSARIAL}
\end{equation}
where $\alpha$ is a hyper-parameter that serves as a learning rate and can be varied at each step, $\nabla_W$ denotes the gradient in the predictor weights $W$ and $proj_{\nabla_W L_A} \nabla_WL$ is a  projection term that prevents the weights from being updated in a direction that benefits the adversary. The last term is the adversary loss which the classifier will now try to minimize.

Another way of creating fair algorithms is through the insertion of a regularizer. This is the idea behind \cite{cite:INPIREG}, where they train a logistic regression model with parameters $\theta$ based on the usual principle of maximizing the log-likelihood:
\begin{equation}
    L(\theta) = \sum_{(y_i, x_i, a_i) \in \mathcal{D}} \log \hat{\mathbb{P}}_{\theta} (y_i \mid x_i, a_i) = \sum_{(y_i, x_i, a_i) \in \mathcal{D}} \log f(y_i \mid x_i, a_i),
    \label{LOSS}
\end{equation}
where the posterior probabilities are obtained through the logistic model.\\
Fairness is incorporated by adding a regularizer that measures the mutual information of the sensitive attribute and the response through a function, $PI$, based on the Prejudice Index,
\begin{equation}
PI = \sum_{
(y,a) \in D
} \mathbb{P}(y,s) \log \frac{
\mathbb{P}(y,s)}{{\mathbb{P}(y)\mathbb{P}(s)}
},
\end{equation}

\noindent which can be approximated by the expression:

\begin{equation}
\widehat{PI} = \sum_{
(x_i, a_i) \in D
} \sum_{
 y \in \{0, 1\}
} f(y \mid x_i, a_i) \log \frac{\hat{\mathbb{P}}(y \mid a)}{\hat{\mathbb{P}}(y)}.
\label{PIREG}
\end{equation}

The insertion of $PI$ as a regularization term in the loss of the classifier penalizes shared information between the predictions and the sensitive attributes. Its influence on the loss can be controlled with a parameter $\eta$ depending on the desired level of fairness. The final loss function by adding the loss function \eqref{LOSS}, the regularizer given by \eqref{PIREG} and the usual $L^2$ penalization:
\begin{equation}
    \mathcal{L}(\theta) = -L(\theta) + \eta \widehat{PI} + \frac{\lambda}{2} || \theta||^2. 
\end{equation}

Finally, we will consider the meta-algorithm in \cite{cite:INMETAFAIR}. Basically, it proposes the inclusion of a fairness constraint to the problem of minimizing the error rate of a classifier. The idea is to express fairness through a measure of group performance, $q_a$ for $a\in A$. This group performance should be written as a probability, $q_a = \mathbb{P}(\xi \mid A = a, \xi')$, where $\xi$, $\xi'$ are events. Then, a certain $\tau$-rule is imposed by incorporating the minimum quotient of group performances in the optimization problem
\begin{equation}
\min \mathbb{P}(\hat{Y} \neq Y ) \quad 
\text{s.t } \min_{a\in A} q_a / \max_{a\in A} q_a \geq \tau.
\label{METAFAIR}
\end{equation}
The choice of $\xi$ and $\xi'$ determines the fairness metric that we implement. A few examples:
\begin{itemize}
    \item If $\xi = \{\hat{Y} = 1\}$, $\xi' = \emptyset$ then $q_a$ will measure independence. 
    \item If $\xi = \{\hat{Y} = 1\}$, $\xi' = \{ Y = 0 \}$ then $q_a$ will measure equal opportunity.
    \item If $\xi_1 = \{\hat{Y} = 1\}$, $\xi_1' = \{ Y = 0 \}$ and $\xi_2 = \{\hat{Y} = 0\}$, $\xi_2' = \{ Y = 1 \}$ then $q^i_a$ will measure separation.
    \item If $\xi = \{Y = 1\}$, $\xi' = \{ \hat{Y} = 1 \}$ then $q_a$ will measure sufficiency. 
\end{itemize}

\subsubsection{Post-Processing}
\label{subsec:post}

Post-processors act once the classification has been done, changing the predicted labels in order to achieve fairness. This is the case of the reject option classifier proposed in \cite{cite:POSTOPTREJECT}. This method relabels predictions with posterior probabilities close to $0.5$ (i.e. the less reliable ones), assigning the positive label to unprivileged individuals and the negative label to the privileged ones. Formally, the relabeling process is controlled by a parameter $\theta$ that determines a band around $0.5$ in such a way that the relabeling is applied only to observations whose posterior probabilities lie in that band. The band is defined as
\begin{equation}
\{ x \in X \mid max [\mathbb{P}(\hat{Y} = 1 \mid x), 1 - \mathbb{P}(\hat{Y} = 1 \mid x)] < \theta \}.
\label{OPTREJECT}
\end{equation}
The fact that the only relabeled individuals are those close to the decision boundary minimizes the incurred error.

Another post-processor is the equal odds processor, described in \cite{cite:POSTEQODDS}. This method proposes to construct a derived classifier, $\widetilde{Y}$, from a given classifier, $\hat{Y}$, or score, $R$. In order to ensure separation, the derived classifier will choose accuracy trade-offs that belong to the intersection of all $a$-conditional ROC curves of the classifier or score. Formally, the authors introduce a series of vectors that store the group $FPR$ and $TPR$ of a classifier $\widehat{Y}$, $  \gamma_a(\widehat{Y}) = ( FPR_{A=a}, TPR_{a=A} )$. Then, they define $D_a$, the convex hull of the image of the $a$-conditional ROC curves. Finally, the derived classifier is the solution to the minimization problem
\begin{equation}
\begin{split}
    \text{
     min
    }& \quad \mathbb{E} [ L(\widetilde{Y}, Y)] \\
    \text{s.t} & \quad \gamma_a(\widetilde{Y}) \in D_a (\widehat{Y  }), \quad \forall a \in A, \\ 
    & \quad \gamma_0(\widetilde{Y}) = \gamma_1(\widetilde{Y}).\\
\end{split}
\label{EQODDS}
\end{equation}

The last post-processor that we consider is one based on the idea of calibration. Calibration by groups is equivalent to sufficiency \cite{cite:FAIRMLBOOK}. One way of obtaining a score from a classification that already has been done is through Platt scaling \cite{cite:POSTPLATT}. Therefore, performing Platt scaling by groups (i.e. restricting the observations to each of the sensitive groups) results in a calibrated score. Platt scaling consists on fitting a logistic regression model to the predictions of a classifier in order to obtain posterior probabilities.

\section{Extending fairness}
\label{sec:extending}

This Section delves into the novel contributions of the paper. First we introduce \emph{logical processors}, a new technique to handle the case of multiple sensitive variables, expanding the domain of application of the processors that have been developed. Then we will study the interaction between different types of processors through \emph
{multistage processors}.

\subsection{Multiple sensitive variables and logical processors}
\label{subsec:multiple}

One of the biggest gaps in fair ML lies in the case of multiple sensitive variables \cite{cite:POSTOPTREJECT,cite:FAIRMLBOOK}. Many of the algorithms that have been proposed in the literature cannot handle multiple sensitive variables or multiple labels. This problematic in certain legal frameworks. For example, financial institutions are forced by law to apply fairness according to multiple criteria at once \cite{caton2024fairness}. Moreover, from an societal point of view it is particularly important to acknowledge and address the prejudice that a given individual may experience due to the combined experience of belonging to multiple sensitive groups. This phenomenon is known as \emph{compounded prejudice} and is well documented in the literature \cite{malloy2010standards}.\\
In order to address the case of multiple sensitive attributes, we propose \emph{logical processors} (LP), which apply a logical function to the sensitive variables in order to codify their information in a single binary variable. Formally, consider a set of sensitive attributes $(A_1, A_2, ..., A_n)$. Then a logical processor is a function $f:(A_1,...,A_n) \longrightarrow \{0,1 \}$. That is, it maps all the attributes into a single binary variable, which allows us to exploit all the tools that have been developed to handle this situation.\\
The simplest example of a LP is to consider the OR sum of two sensitive attributes. The resulting variable indicates the presence of a sensitive attribute in a given individual. However, notice that the operation we use has deep implications. Suppose, for instance, that the sensitive attributes are race and sex. Their OR sum acknowledges that the individual may be a person of color or a woman, but it does not distinguish between the two, which may be a mistake because different minorities may experience different types of discrimination \cite{ro2009social}. Furthermore, it cannot identify women of color, which is problematic due to compounded prejudice. This is why it is necessary to consider many different logical processors and their implications. In particular, we will tackle the bivariate case and consider:

\begin{enumerate}
    \item \textbf{OR}: Presence of one or multiple sensitive variables. An individual is considered unprivileged if they have any of the two sensitive attributes. This approach does not take compounded prejudice into consideration.
    \item \textbf{AND}: Exclusive presence of multiple sensitive variables. An individual is considered unprivileged if they have two sensitive attributes. This is particularly useful to identify and address compounded prejudice, but it fails to take individual bias into account.
    \item \textbf{XOR}: Exclusive presence of a single sensitive variable. An individual is not considered unprivileged if they have multiple sensitive attributes. This approach completely ignores compounded prejudice, but it may help to address individual discrimination.
\end{enumerate}

Most of the fairness processors of Section \ref{sec:processors} can handle multiple sensitive attributes. However, our proposal allows for more flexibility when choosing or designing a fairness method: any univariate idea can be extended to the multivariate case in a straightforward manner. Furthermore, it is much easier to implement a LP than it is to adapt a given algorithm or metric.\\

\subsection{Multistage processors}
\label{subsec:integration}

The combination of fairness methods that attack different stages of the ML pipeline has already been mentioned \cite{cite:multiplemotivation}, but it has not yet been implemented in practise. However, the interaction between different processors may lead to a synergistic interaction that provides improved fairness or accuracy. We introduce and formalize this notion through \emph{multistage processors} (MP), which are defined as a fairness processor that involves multiple stages of the ML pipeline.\\
The above definition allows for great flexibility when designing a MP, but the study will be restricted to the combination of two fairness processors that involve different parts of the algorithm-making process. Concretely, the following types of MP will be considered:
\begin{enumerate}
    \item \textbf{PI}, involving a pre-processor and a in-processor. 
    \item \textbf{PP}, involving a pre-processor and a post-processor.
    \item \textbf{IP}, involving a in-processor and a post-processor.
\end{enumerate}

\section{Experimental setup}
\label{sec:setup}

This Section provides an in-depth explanation of the methodology and experiments followed to explore the properties of logical and multistage processors.
The analysis consists of two parts: first, a simulation study intended to check the capabilities of the proposed ideas in a controlled environment. Then, a case study based on the data set \emph{german} \cite{cite:GERMAN} to  ascertain the properties of the methods in a real world setting. The code used to implement this methodology can be found in a Github\footnote{\url{https://github.com/arturo-perez-peralta/FairMLIntegration}} repository.\\
The data of the simulation study is generated in a similar way to the toy scenario found in \cite{cite:INADVERSARIAL}, although some modifications were necessary to handle multiple sensitive variables. The simulation schema is as follows: Let $a_i^{(j)} \in \{0,1\}$ with $i = 1,2; j = 1,..., N$ be picked uniformly at random. They represent the sensitive variables. Let $v_i^{(j)} \sim \mathcal{N}(a_i^{(j)}, 1)$ be an innacurate measurement of the sensitive variables, $\epsilon_i^{(j)} \sim \mathcal{N}(1/2,1)  $ be two independent and unrelated attributes, and let $v^{(j)} = (v_1^{(j)} + v_2^{(j)} + \epsilon_1^{(j)} + \epsilon_2^{(j)} )/4$ be their mean. Let $u^{(j)}, w^{(j)} \sim \mathcal{N}(v^{(j)},1)$ be two independent variables. $w^{(j)}$ represents an inaccurate measurement of $v^{(j)}$, and its sign defines the response variable: $y^{(j)} = \mathbf{1}(w^{(j)} > 0)$. On the other hand, $u^{(j)}$ is another noisy measurement of $v^{(j)}$ that serves as a proxy to $w^{(j)}$. Finally, the simulated dataset is $D = (X, Y, A)$ with $X = (a_1^{(j)}, a_2^{(j)}, u^{(j)})$, $Y = (\mathbf{1}(w^{(j)} > 0))_{j=1}^n$, $A = (a_1^{(j)}, a_2^{(j)})_{j = 1}^n$. The simulation is performed $50$ times and the data is aggregated using the median.\\
On the other hand, the german data set includes a different number of features that describe each loan applicant. The target variable $Y$ is binary and represents whether the individual repays the loan or defaults, hence classifying them as good ($Y = 1$) or bad ($Y = 0$) credit risk. In the univariate case we followed \cite{cite:kamiran2009classifying} and used age as a sensitive variable, assuming privilege on those whose age was greater than $25$ and discrimination against those younger. When handling two sensitive attributes we also used sex, assuming discrimination against women.\\
In both data sets two scenarios were considered depending on the use of a single or multiple sensitive variables. Moreover, the multivariate case was divided into three cases depending on which one of the three LP was used (OR, AND, XOR). Therefore, there are four different cases to account for: single sensitive variable, OR, AND, XOR. The relevant characteristics for each case are summarized in Table \ref{tab:datasets}.\\

\begin{table}[h!]
    \centering
    \begin{tabular}{c c c c c c c c c}
         \makecell{Data \\set} & \makecell{Sample\\ size} & Features & \makecell{Default\\ rate} & \makecell{$A_1$\\ rate}  & \makecell{$A_2$\\ rate} & \makecell{OR\\ rate} & \makecell{AND\\ rate} & \makecell{XOR\\ rate} \\
         \hline
         Simulation & $5000$ & $6$ & $0.17$ & $0.50$ & $0.50$ & $0.75$ & $0.25$ & $0.50$\\
         German & $1000$ & $61$ & $0.30$ & $0.15$ & $0.19$ & $0.39$ & $0.11$ & $0.29$
    \end{tabular}
    \caption{Data sets used for credit scoring. $A_1$ and $A_2$ are understood as the first and second sensitive variables. In the German data set, they represent age and sex, respectively.}
    \label{tab:datasets}
\end{table}

The data is split into train, validation and test sets. The models learn a score on the training set, while the threshold is selected to maximize the balanced accuracy \eqref{balacc} in validation. Finally, the metrics seen in Section \ref{sec:background} are evaluated in test, with emphasis on accuracy as a measure of performance and separation as a measure of fairness. When dealing with multiple sensitive variables, the fairness metrics are computed with respect to the same sensitive variable as in the univariate case, with the idea of checking how the use of LP exacerbates or diminishes discrimination on that very same variable.\\
The eight fairness processors seen in Section \ref{sec:processors} were implemented with set values for the hyper-parameters. Afterwards, the MPs consisting of all the combinations seen in Section \ref{subsec:integration} were trained without changing the hyper-parameters, the idea being to observe the interactions between methods as they were. In both cases pre-processors and post-processors were trained with a logistic regression model unless they were part of a MP that also involved an in-processor.\\

\subsection{Simulation study}
\label{sec:Simulation}

\begin{figure}[h!]
        \centering
        \begin{subfigure}[b]{0.49\textwidth}
            \centering
            \includegraphics[width=\textwidth]{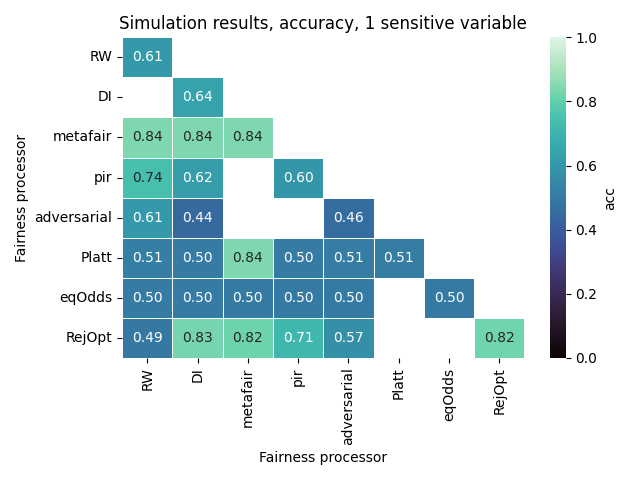}
            \label{fig:matrixSimul1VaccMas}
        \end{subfigure}
        \hfill
        \begin{subfigure}[b]{0.49\textwidth}  
            \centering 
            \includegraphics[width=\textwidth]{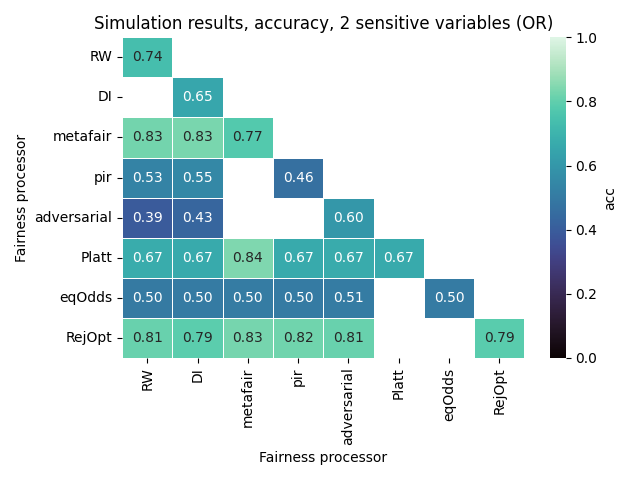}
            \label{fig:matrixSimul2VORaccMas}
        \end{subfigure}
        
        \begin{subfigure}[b]{0.49\textwidth}   
            \centering 
            \includegraphics[width=\textwidth]{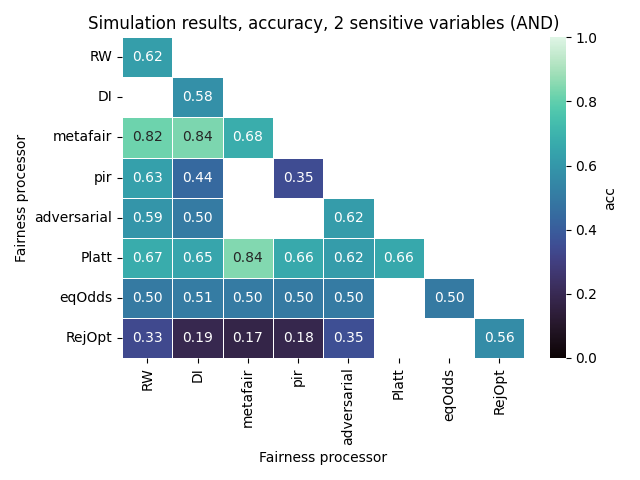}
            \label{fig:matrixSimul2VANDaccMas}
        \end{subfigure}
        \hfill
        \begin{subfigure}[b]{0.49\textwidth}   
            \centering 
            \includegraphics[width=\textwidth]{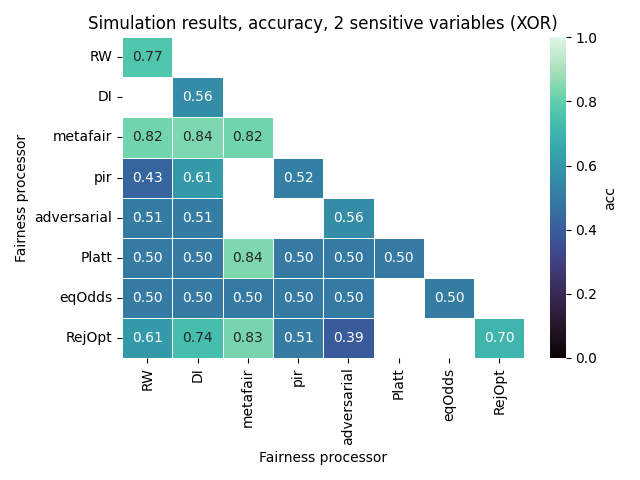}
            \label{fig:matrixSimul2VXORaccMas}
        \end{subfigure}
        \caption[Accuracy of multistage processors in the simulation study. The main diagonal shows the accuracy of each method while the off-diagonal elements represent the performance of the combination of two processors.]
        {\small Accuracy of multistage processors in the simulation study. The main diagonal shows the accuracy of each method while the off-diagonal elements represent the performance of the combination of two processors.} 
        \label{fig:AccuracySimulMas}
    \end{figure}

\begin{figure}[h!]
        \centering
        \begin{subfigure}[b]{0.49\textwidth}
            \centering
             \includegraphics[width=\textwidth]{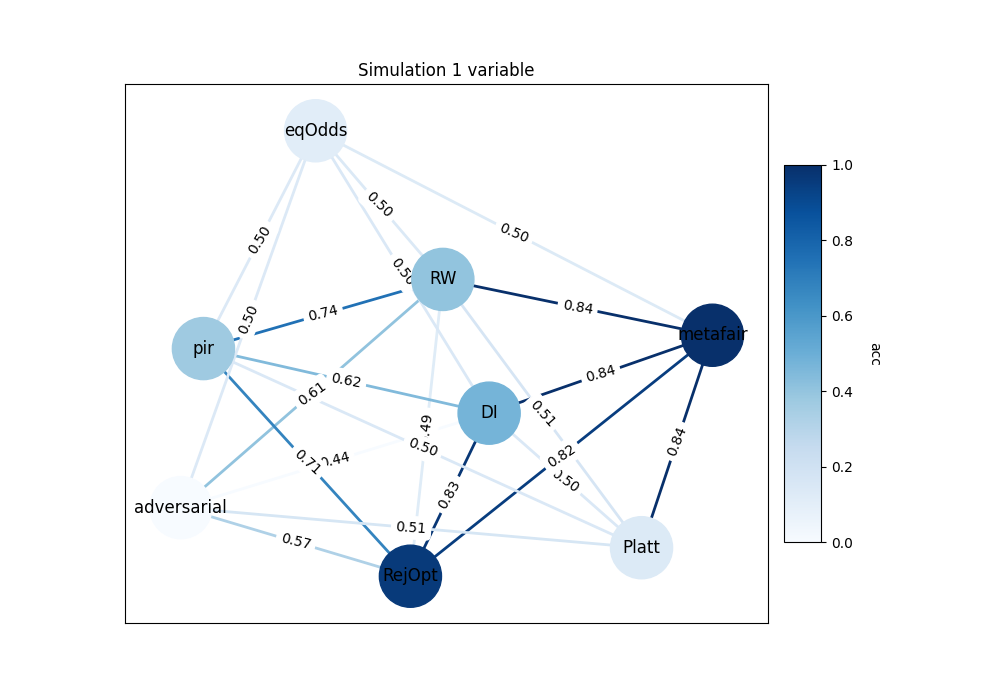}
            \label{fig:graphSimul1VaccMas}
        \end{subfigure}
        \hfill
        \begin{subfigure}[b]{0.49\textwidth}  
            \centering 
            \includegraphics[width=\textwidth]{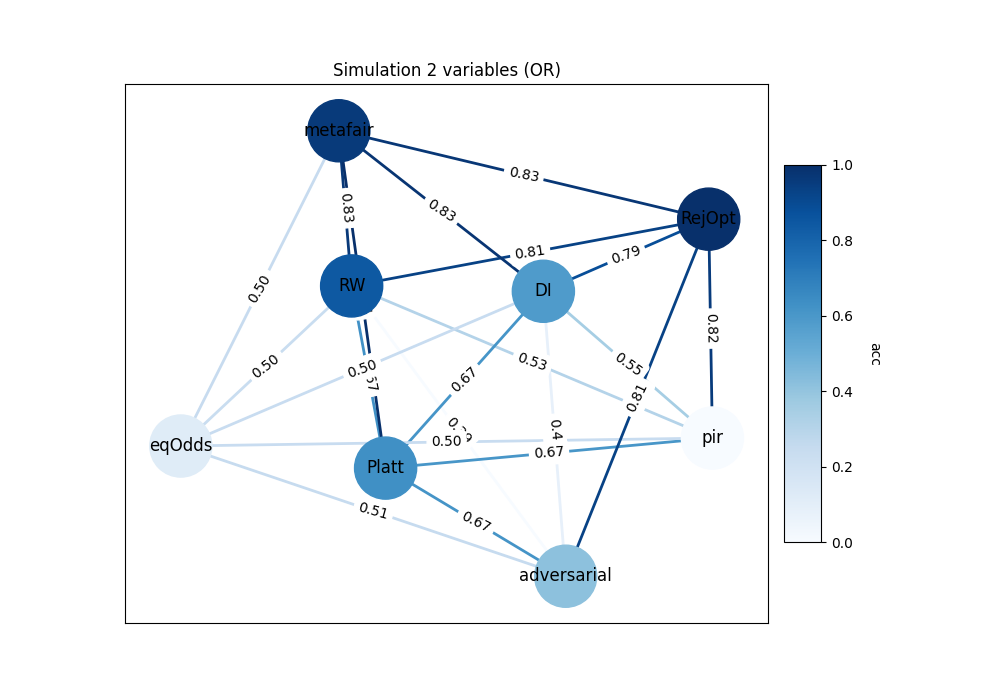}
            \label{fig:graphSimul2VORaccMas}
        \end{subfigure}
        \begin{subfigure}[b]{0.49\textwidth}   
            \centering 
            \includegraphics[width=\textwidth]{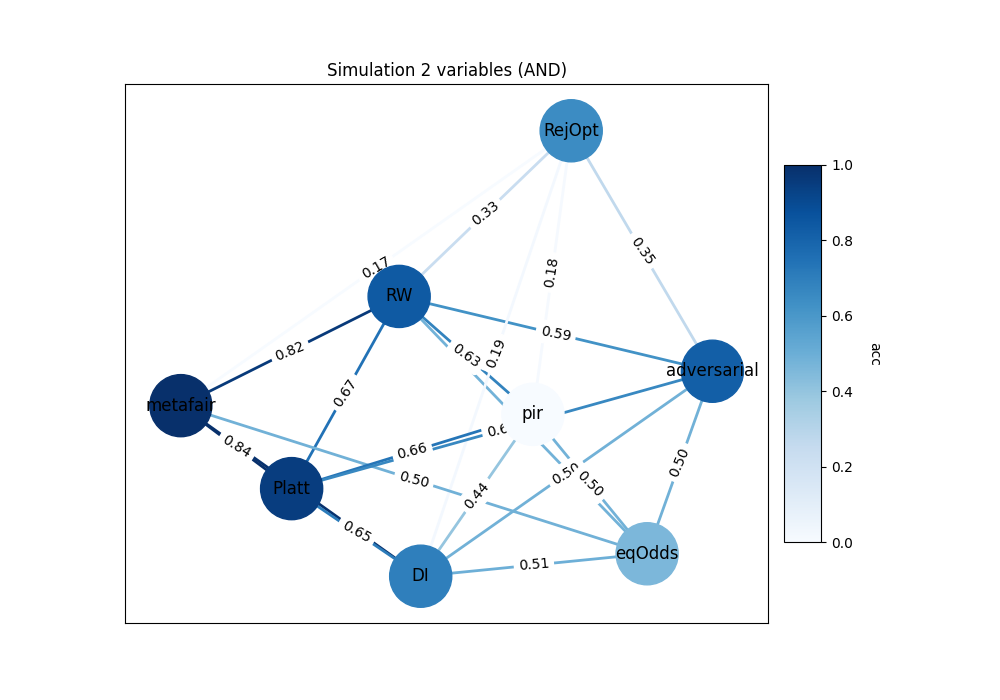}
            \label{fig:graphSimul2VANDaccMas}
        \end{subfigure}
        \hfill
        \begin{subfigure}[b]{0.49\textwidth}   
            \centering 
            \includegraphics[width=\textwidth]{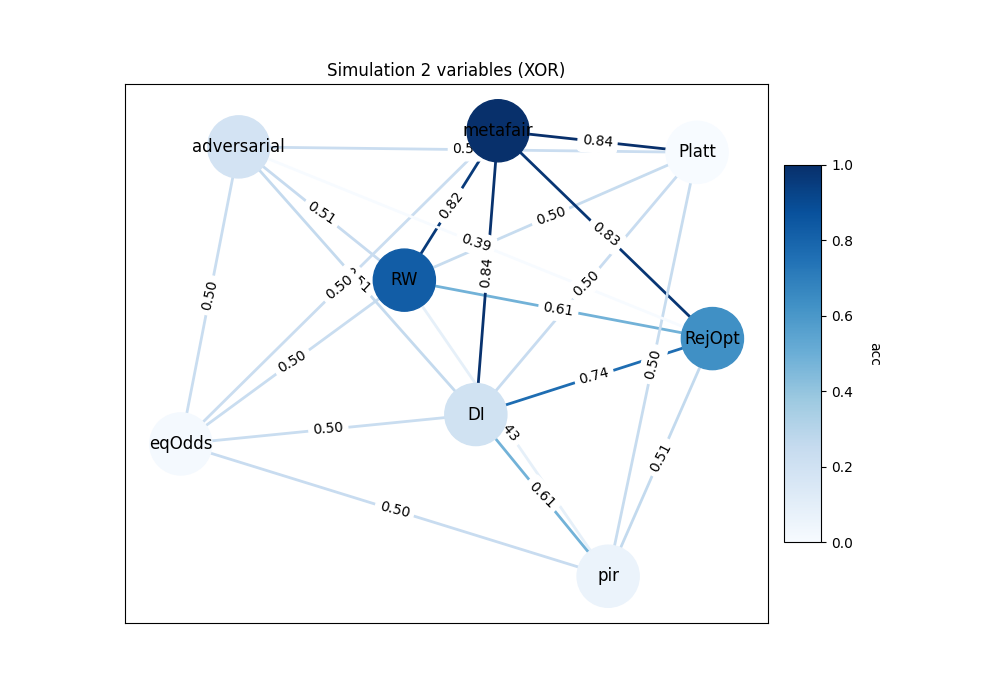}
            \label{fig:graphSimul2VXORaccMas}
        \end{subfigure}
        \caption[Accuracy graph of multistage processors for the simulation study. Nodes show the accuracy of individual methods while edges represent the performance of the combination of two methods.]
        {\small Accuracy graph of multistage processors for the simulation study. Nodes show the accuracy of individual methods while edges represent the performance of the combination of two methods.} 
        \label{fig:graphAccSimulMas}
    \end{figure}

\begin{figure}[h!]
    \centering
        \begin{subfigure}[b]{0.49\textwidth}
            \centering
            \includegraphics[width=\textwidth]{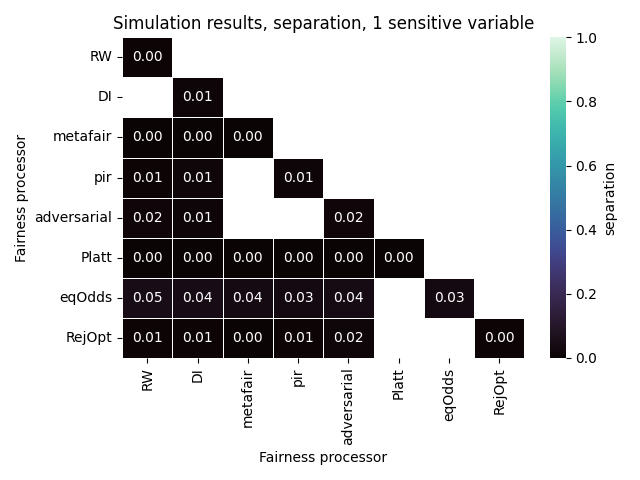}
            \label{fig:matrixSimul1VsepMas}
        \end{subfigure}
        \hfill
        \begin{subfigure}[b]{0.49\textwidth}  
            \centering 
            \includegraphics[width=\textwidth]{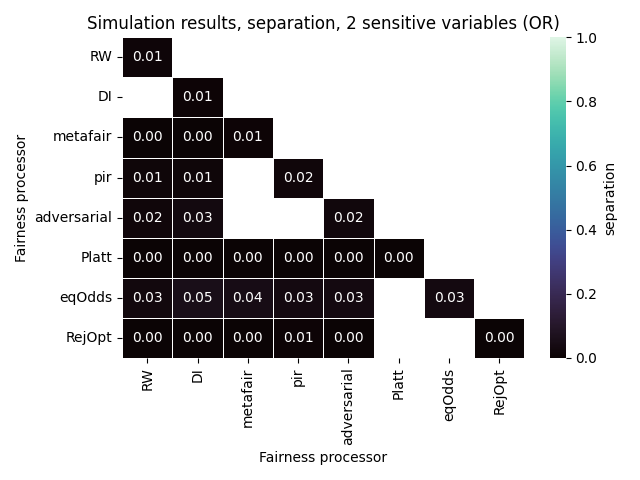}
            \label{fig:matrixSimul2VORsepMas}
        \end{subfigure}
        \vskip\baselineskip
        \begin{subfigure}[b]{0.49\textwidth}   
            \centering 
            \includegraphics[width=\textwidth]{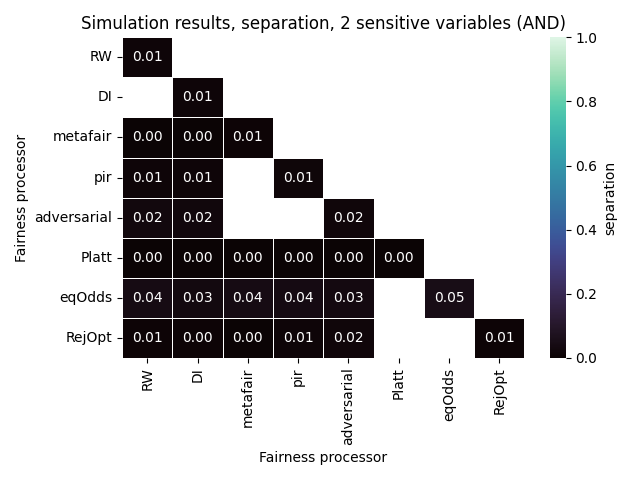}
            \label{fig:matrixSimul2VANDsepMas}
        \end{subfigure}
        \hfill
        \begin{subfigure}[b]{0.49\textwidth}   
            \centering 
            \includegraphics[width=\textwidth]{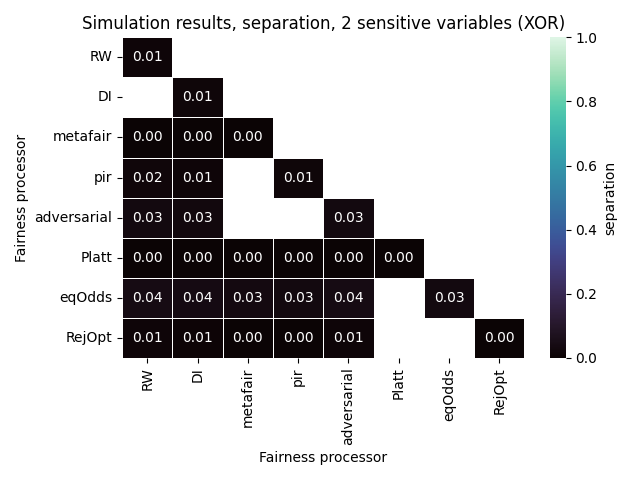}
            \label{fig:matrixSimul2VXORsepMas}
        \end{subfigure}
        \caption[Separation heatmap for the simulation study.]
        {\small Separation heatmap for the simulation study.} 
        \label{fig:SepSimulMas}
    \end{figure}

\begin{figure}[h!]
        \centering
        \begin{subfigure}[b]{0.49\textwidth}
            \centering
            \includegraphics[width=\textwidth]{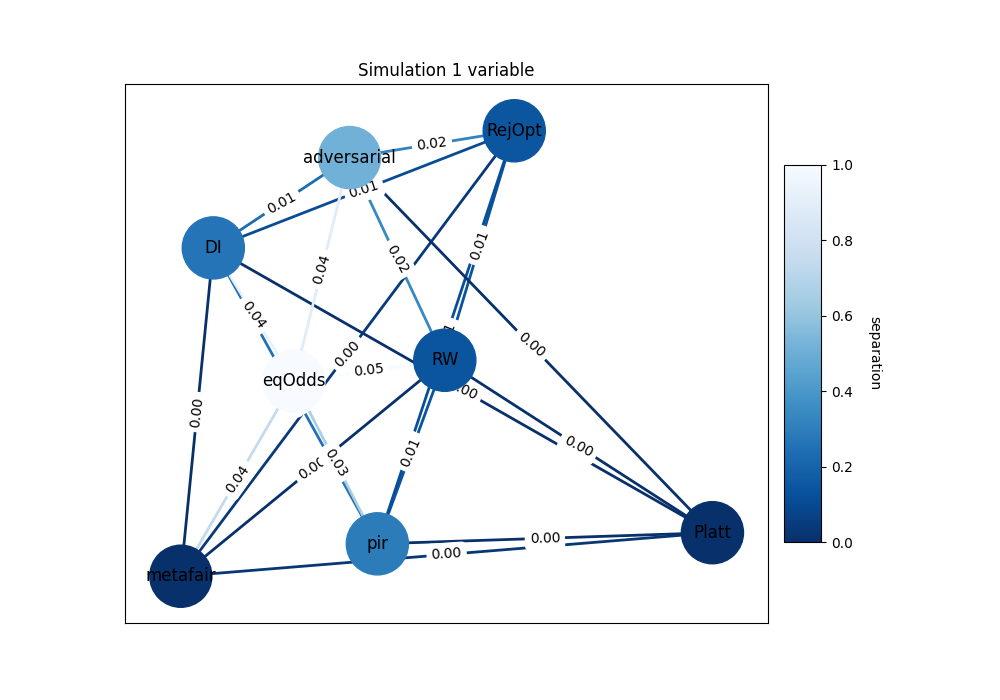}
            \label{fig:graphSimul1VsepMas}
        \end{subfigure}
        \hfill
        \begin{subfigure}[b]{0.49\textwidth}  
            \centering 
            \includegraphics[width=\textwidth]{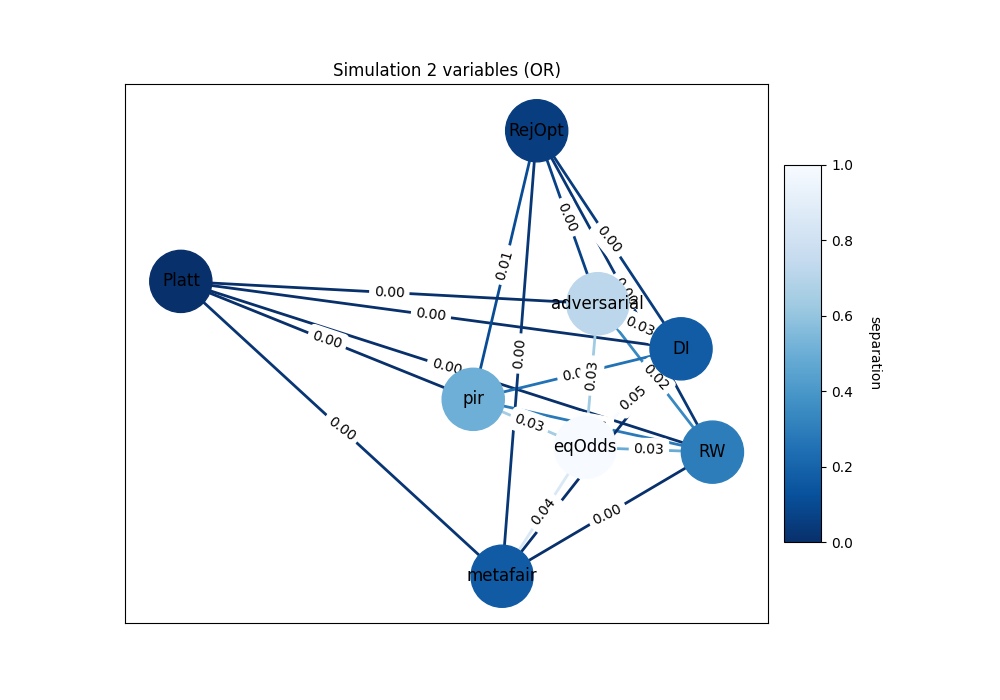}
            \label{fig:graphSimul2VORsepMas}
        \end{subfigure}
        \vskip\baselineskip
        \begin{subfigure}[b]{0.49\textwidth}   
            \centering 
            \includegraphics[width=\textwidth]{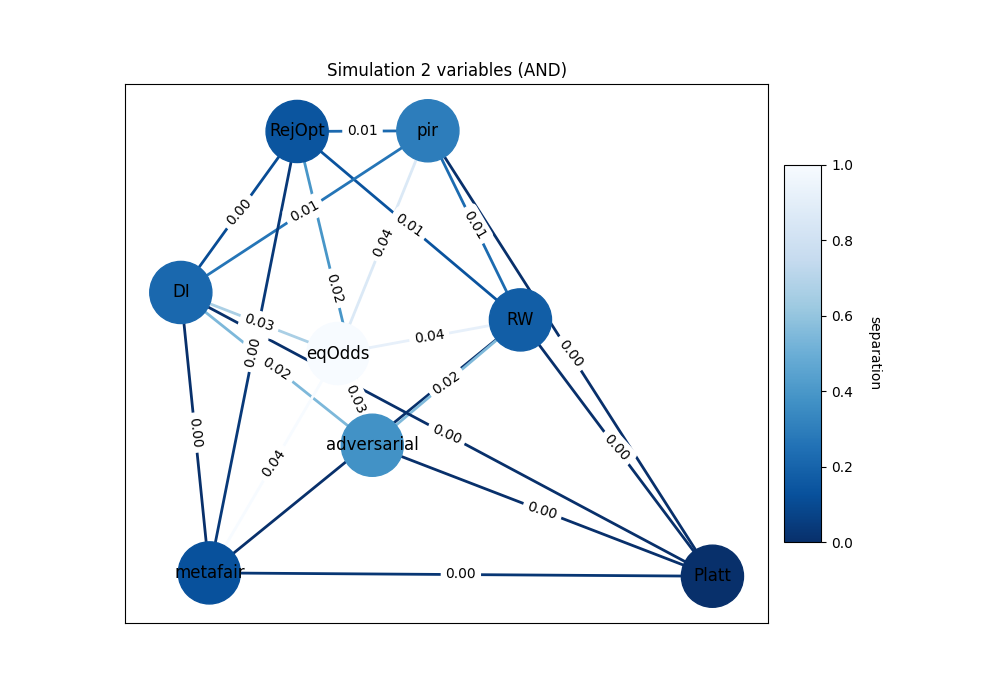}
            \label{fig:graphSimul2VANDsepMas}
        \end{subfigure}
        \hfill
        \begin{subfigure}[b]{0.49\textwidth}   
            \centering 
            \includegraphics[width=\textwidth]{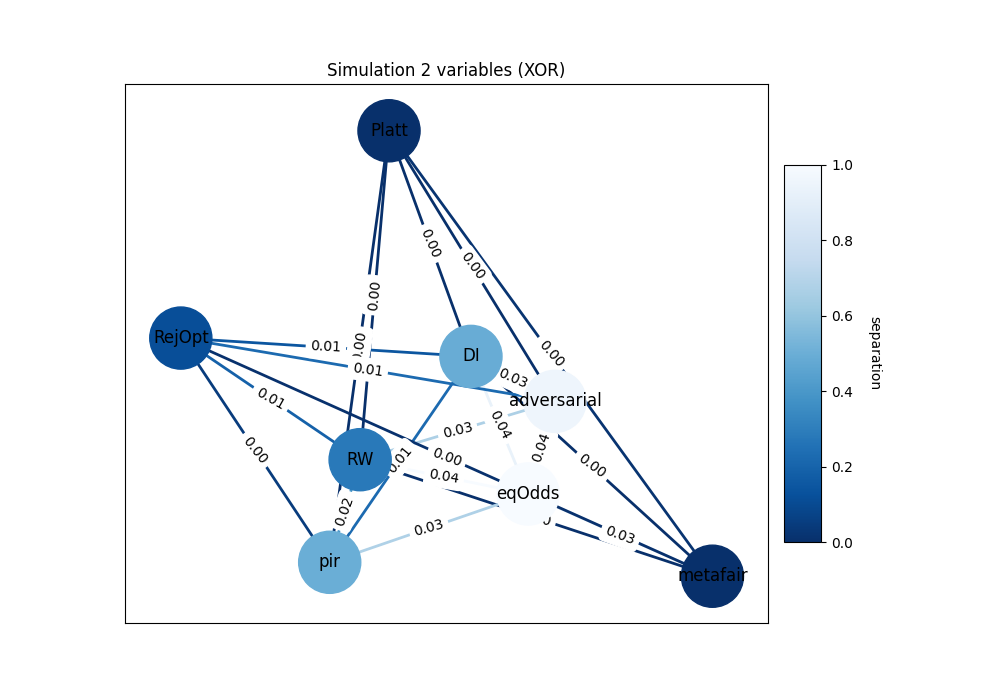}
            \label{fig:graphSimul2VXORsepMas}
        \end{subfigure}
        \caption[Separation graph for the simulation study]
        {\small Separation graph for the simulation study} 
        \label{fig:graphSepSimulMas}
    \end{figure}

\begin{figure}[h!]
        \centering
        \begin{subfigure}[b]{0.49\textwidth}
            \centering
            \includegraphics[width=\textwidth]{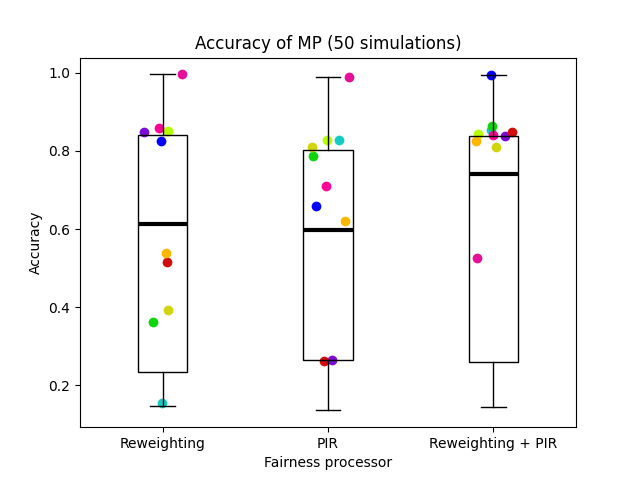}
            \label{fig:BoxplotsMP}
        \end{subfigure}
        \hfill
        \begin{subfigure}[b]{0.49\textwidth}  
            \centering 
            \includegraphics[width=\textwidth]{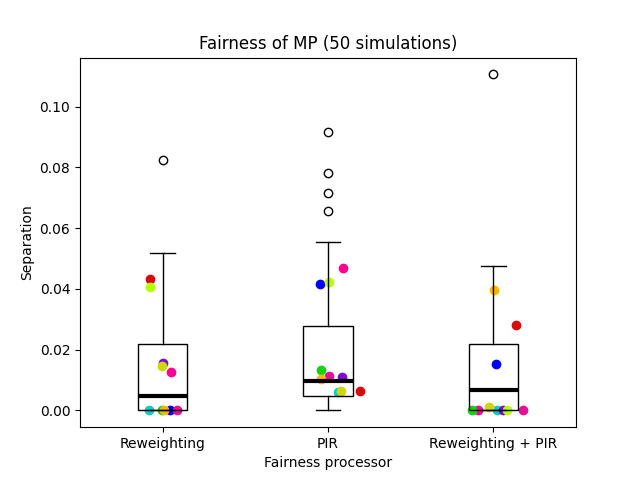}
            \label{fig:BoxplotsFairMP}
        \end{subfigure}
        \vskip\baselineskip
        \begin{subfigure}[b]{0.49\textwidth}   
            \centering 
            \includegraphics[width=\textwidth]{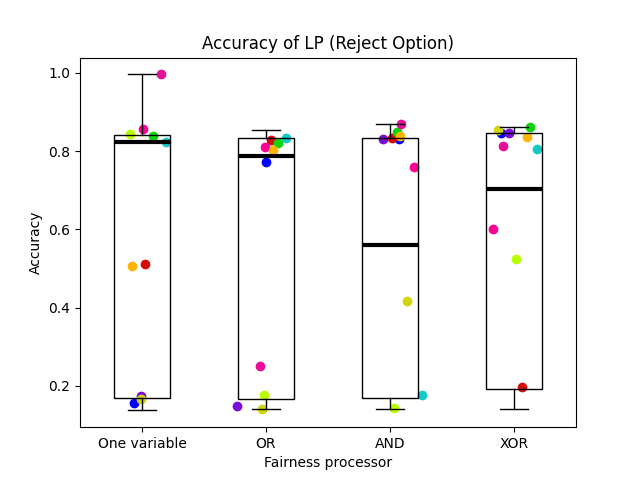}
            \label{fig:BoxplotsAccLP}
        \end{subfigure}
        \hfill
        \begin{subfigure}[b]{0.49\textwidth}   
            \centering 
            \includegraphics[width=\textwidth]{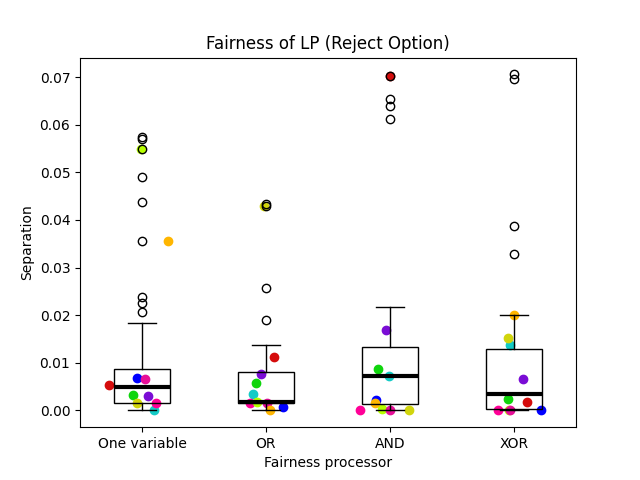}
            \label{fig:BoxplotsFairLP}
        \end{subfigure}
        \caption[Boxplots of the results of the simulation study for selected methods. Ten individual instances selected at random are highlighted with different color for easier tracing.]
        {\small Boxplots of the results of the simulation study for selected methods. Ten individual instances selected are random are highlighted with different color for easier tracing.} 
        \label{fig:Boxplots}
    \end{figure}

The results for the simulation study are summarized in Figures \ref{fig:AccuracySimulMas} to \ref{fig:Boxplots}. We start the analysis by checking the median accuracy of the methods for $50$ simulations, which can be visualized in Figures \ref{fig:AccuracySimulMas} and \ref{fig:graphAccSimulMas}. 

At first glance, the simulation shows a moderate success: the combined use of fairness processors can lead to improvements in performance. This is the case of the MP consisting of reweighting and prejudice index regularizer. These two processors have a median accuracy of around $0.6$ in the  univariate case. However, the MP that combines them reaches an accuracy level of $0.74$; that is, an improvement of around $20\%$. These results suggest that the interaction between fairness processors can lead to improved performance, which is promising in and of itself. Moreover, they can be used to achieve top performance in a data set: this is the case both in the OR and AND processing scenarios, where no individual method can achieve the highest accuracies. In more general terms, the performance of a MP is not bounded above or below by the accuracy of its constituents. We just saw an example of the former, for the later we can take reweighting ($0.61$), reject option ($0.82$), and their combination ($0.49$). However, MPs with lower accuracy than their constituents seem to be a minority, and the combination of two methods can be expected to perform at worst like the worse of its two constituents. 

PI combinations seem to yield the most stable results overall both in the univariate and multivariate cases. The interaction of pre-processors and in-processors yield the best performance when using the metafair classifier, while methods that involve the prejudice index regularizer or adversarial debiasing are a mixed bag with their accuracy depending on the choice of LP. The processor that acts last (in-processor in PI combinations and post-processors in both PP and IP MPs) seems to have a great impact on the performance of the resulting MP, exhibiting a kind of dominating behavior. This phenomenon is particularly striking when using a post-processor, specially with Platt scaling in which all methods that involve it have an accuracy of around $0.5$. The use of Reject option can lead to highly variable results, reaching accuracy levels as low as $0.17$ when using AND processsing and as high as $0.83$ in the OR scenario. In general, IP processors are able to achieve better performance than PP processors. 

The choice of logical processor can be a great cause of variation for certain processors. For instance, reject option seems to work best when using the OR procedure, with its performance being slightly compromised when considering XOR and it becoming abysmal if we opt for AND processing. Other processors, such as those combinations involving the metafair classifier, seem quite stable, reaching similar accuracy in all scenarios. In any case, the choice of LP is not trivial and should be studied carefuly along with each considered processor.


Now we proceed with the analysis of separation with the aid of Figures \ref{fig:SepSimulMas} and \ref{fig:graphSepSimulMas}. 
In general, the results are very uniform, reaching a level of separation close to $0$ in all instances. Like accuracy, separation of a MP is not bounded below (take the combination $0.00$ of disparate impact remover $0.01$ and the metafair classifier $0.01$ in the univariate case) or above (take the combination $0.01$ of reweighting $0.00$ and reject option $0.01$ in the univariate case) by the fairness of its constituents. In any case, the use of MPs can also lead to improved fairness. In particular, some MPs seem to be particularly promising: take the combination of the disparate impact remover and the metafair classifier when using OR processsing. Table \ref{tab:poc} shows how their interaction leads to one of the best solutions, being a top performer both in terms of accuracy (it presents the best accuracy when using OR processing) and fairness (achieving perfect separation) even though their constituents show worse accuracy.

\begin{table}[h!]
    \centering
    \begin{tabular}{c | c c c}
        Processor & Accuracy & Fairness  \\
        \hline
        Disparate impact remover & $0.66$ & $0.01$ \\
        Metafair classifier & $0.77$ & $0.08$ \\
        Combination & $0.83$ &  $0.00$
    \end{tabular}
    \caption{Median performance and fairness metrics for the combination of reweighting and prejudice index regularizer in the univariate case.}
    \label{tab:poc}
\end{table}

Finally, in Figure \ref{fig:Boxplots} we study the effect of MPs and LPs in more detail through boxplots that allow us to analyze the distribution of both accuracy and separation. Furthermore, in order to better study the correlations between different methods, we highlight ten individual simulation instances selected at random with different color. To study MPs we turn our attention to the combination of reweighting with the prejudice index regularizer which yielded good results in terms of both fairness and accuracy. To explore LPs we use the reject option classifier which was particularly sensitive to changes in LP.

The results for reweighting and the prejudice index regularizer show that those classifiers do not perform on average as badly as they did on the simulation instance we considered, nor does their combination outperform them so severely as it did. Nonetheless, the fact remains that their combination yields a significant improvement in accuracy. If we turn our attention to individual points we can find some instances in which the use of a MP yields significant improvement over individual fairness processors. This is particularly remarkable in the case of the red point, which produces a very accurate classifier in a setting in which both individual processors perform relatively poorly, and the blue point, in which two accurate algorithms achieve perfect accuracy when combined. In general, the improvement is noticeable but moderate, and in the worst case scenario the use of a MP can lead to a sharp decrease in accuracy, which is the case of the pink dot. In terms of fairness the use of just reweighting produces results that are slightly more fair in terms of separation, but the improvement is not that significant. Upon closer inspection, most simulation instances lie close to zero for all three methods. It is particularly remarkable the case of solutions like the greener ones which are relatively unfair and inaccurate when separate but perfectly fair and top-performing when combined. However, there are also cases like the orange dot which produces a relatively unfair solution from fair constituents. All in all, when looking at the scale of the graph it is clear that all solutions are close to perfect separation. Overall, the combination of these two methods does seem to produce a better, fair classifier.

Turning our attention to the results for reject option we can see that performance is consistently degraded when considering multiple sensitive variables, although this effect depends on the choice of LP: OR processing barely compromises accuracy, XOR produces a reduction in performance of around $12\%$ and the use of AND produces a loss of nearly $25\%$ in accuracy. On the other hand, note that the use of LPs seem to yield lower dispersion for the distribution of accuracy. When looking at individual solutions it is particularly striking how the use of some logical processors can compromise the accuracy of certain classifiers (take, for instance, the pink point with OR processing or the greenish dot with XOR). However, it is remarkable how some instances achieve leaps in accuracy by simply using logical processing. This is the case for the yellow point, which showed terrible performance (around $0.2$) in the univariate case reaches an increase of accuracy of nearly $400\%$ when using XOR processing, managing to improve its accuracy to $0.8$; and for the blue dot which is similarly upgrading when using OR processing. In any case, the overall tendency is for individual solutions to compromise their performance in at least one of the cases. In terms of fairness the differences, while not as acute, are still present. Both OR and XOR yield better results in terms of fairness than the case of a single sensitive variable, with the former providing the lowest separation; the use of AND, on the other hand, yields a worse classifier in terms of fairness. When looking at individual instances we can see how certain logical processors can hamper separation. This is the case for the red point, whose fairness is compromised to the point of becoming an outlier when using AND processing. On the other hand, instances like the orange one find big improvements under any kind of logical processing. All in all, the most appropriate tool for handling multiple sensitive variables for reject option seems to be OR processing. 

This simulation study serves as a satisfactory demonstration that logical and multistage processors can preserve both fairness and accuracy and, furthermore, improve them. However, the conclusions we have drawn are a result of the aggregation of multiple simulation instances. In order to better grasp the performance of the different methods in an individual setting we have included the results for a single simulation instance in the Appendix. All in all, the results of the simulations are very optimistic and have provided useful insights that will be put to the test in the case study.

\subsection{Case study: German data set}

\begin{figure}[h!]
        \centering
        \begin{subfigure}[b]{0.49\textwidth}
            \centering
            \includegraphics[width=\textwidth]{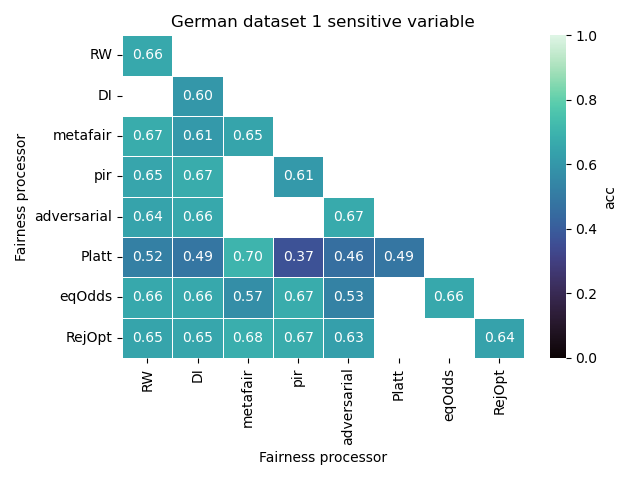}
            \label{fig:matrixGerman1Vacc}
        \end{subfigure}
        \hfill
        \begin{subfigure}[b]{0.49\textwidth}  
            \centering 
            \includegraphics[width=\textwidth]{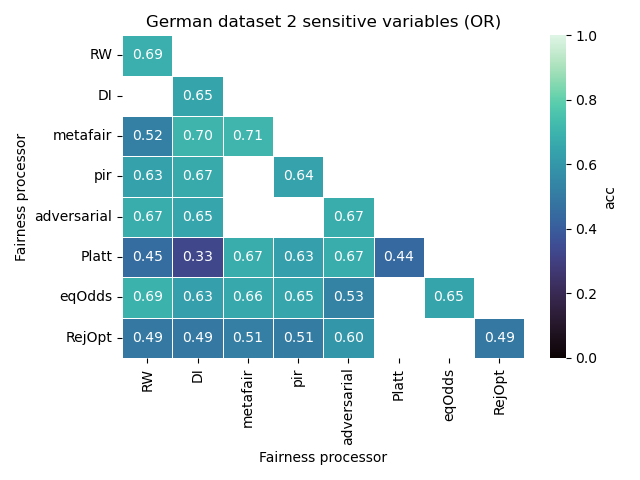}
            \label{fig:matrixGerman2VORacc}
        \end{subfigure}
        \vskip\baselineskip
        \begin{subfigure}[b]{0.49\textwidth}   
            \centering 
            \includegraphics[width=\textwidth]{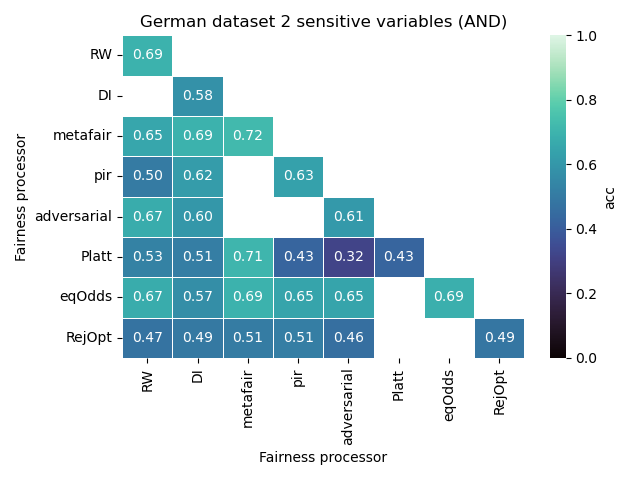}
            \label{fig:matrixGerman2VANDacc}
        \end{subfigure}
        \hfill
        \begin{subfigure}[b]{0.49\textwidth}   
            \centering 
            \includegraphics[width=\textwidth]{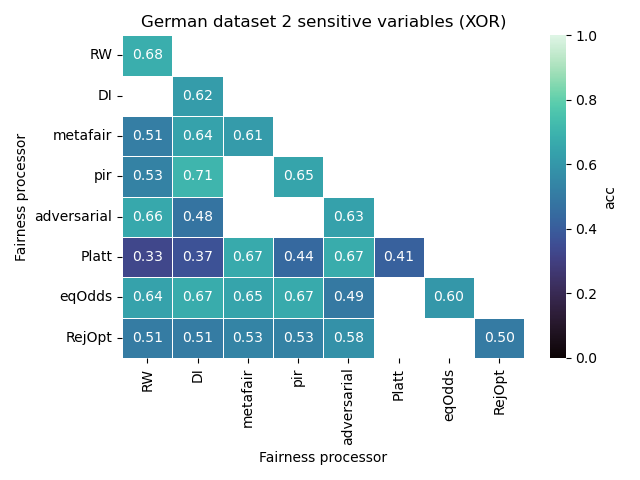}
            \label{fig:matrixGerman2VXORacc}
        \end{subfigure}
        \caption[Accuracy heatmap for the German data set]
        {\small Accuracy heatmap for the German data set} 
        \label{fig:AccuracyGerman}
    \end{figure}

\begin{figure}[h!]
        \centering
        \begin{subfigure}[b]{0.49\textwidth}
            \centering
            \includegraphics[width=\textwidth]{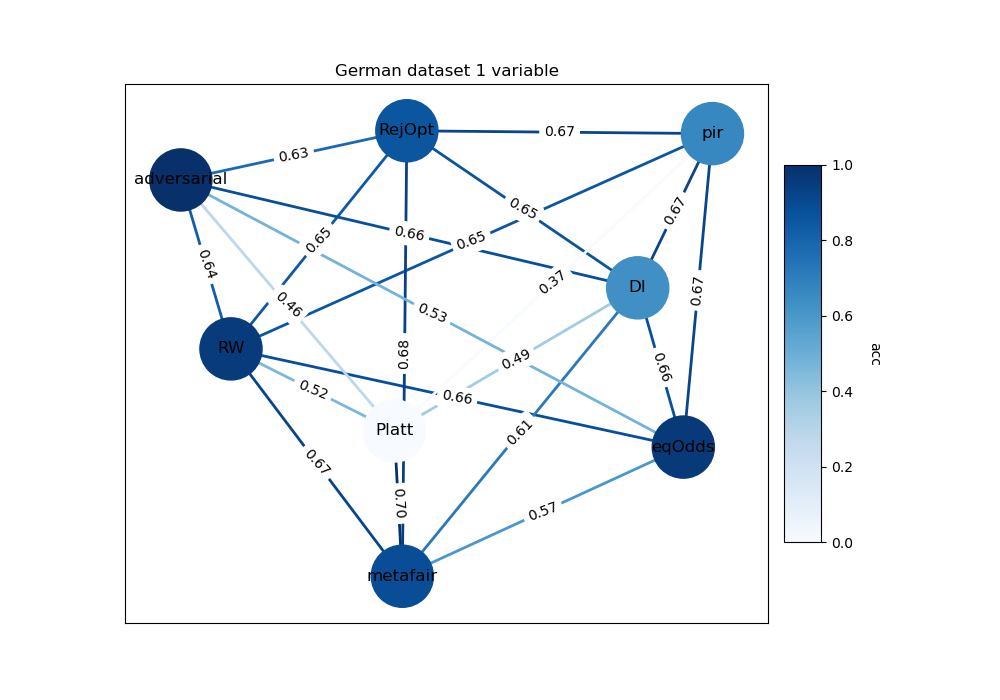}
            \label{fig:graphGerman1Vacc}
        \end{subfigure}
        \hfill
        \begin{subfigure}[b]{0.49\textwidth}  
            \centering 
            \includegraphics[width=\textwidth]{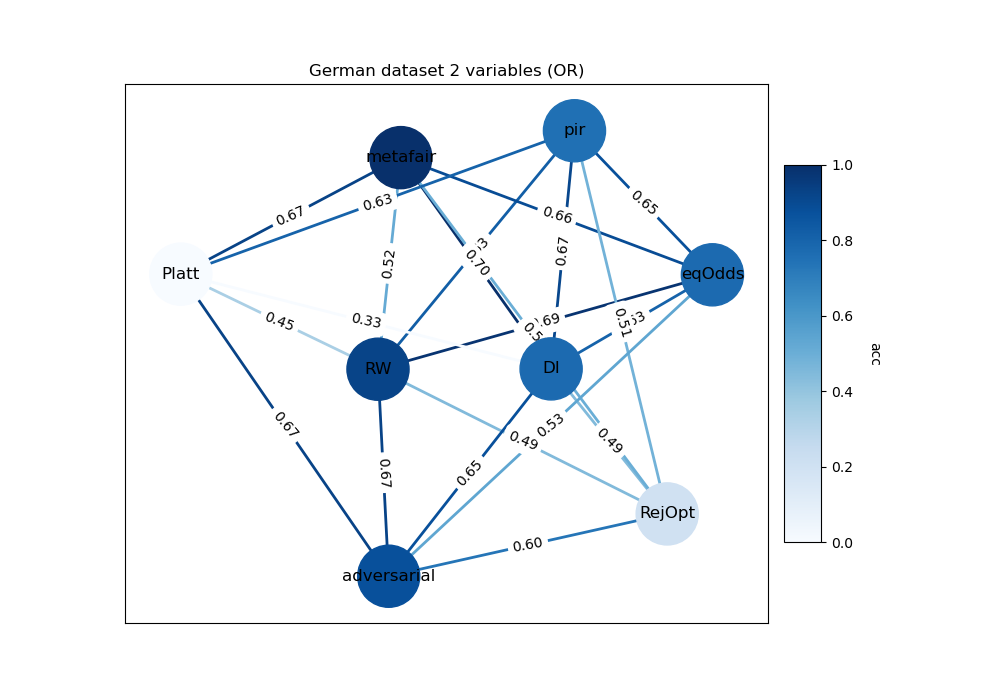}
            \label{fig:graphGerman2VORacc}
        \end{subfigure}
        \vskip\baselineskip
        \begin{subfigure}[b]{0.49\textwidth}   
            \centering 
            \includegraphics[width=\textwidth]{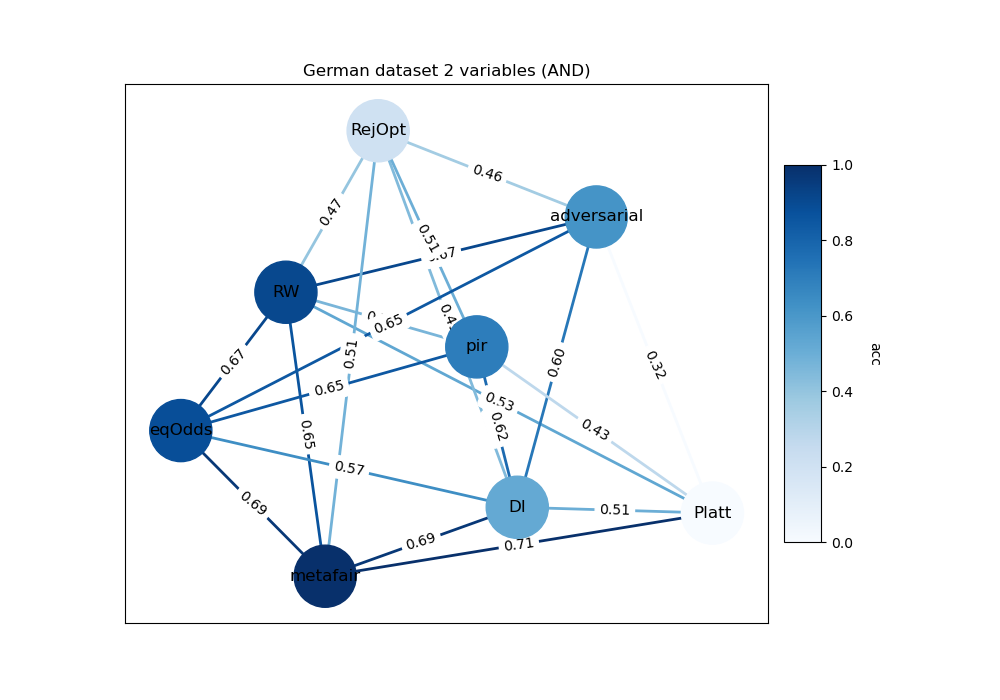}
            \label{fig:graphGerman2VANDacc}
        \end{subfigure}
        \hfill
        \begin{subfigure}[b]{0.49\textwidth}   
            \centering 
            \includegraphics[width=\textwidth]{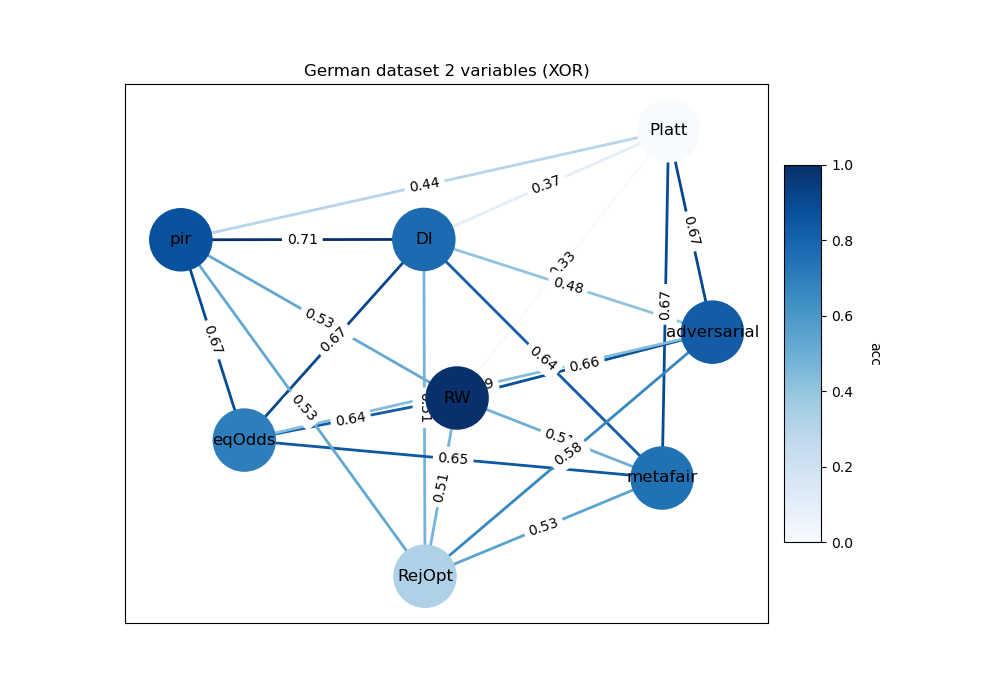}
            \label{fig:graphGerman2VXORacc}
        \end{subfigure}
        \caption[Accuracy graph in German data set]
        {\small Accuracy graph in German data set} 
        \label{fig:graphAccGerman}
    \end{figure}

\begin{figure}[h!]
        \centering
        \begin{subfigure}[b]{0.49\textwidth}
            \centering
            \includegraphics[width=\textwidth]{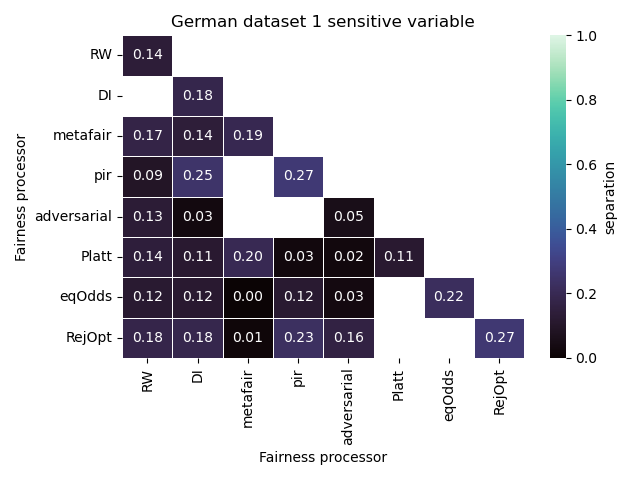}
            \label{fig:matrixGerman1Vsep}
        \end{subfigure}
        \hfill
        \begin{subfigure}[b]{0.49\textwidth}  
            \centering 
            \includegraphics[width=\textwidth]{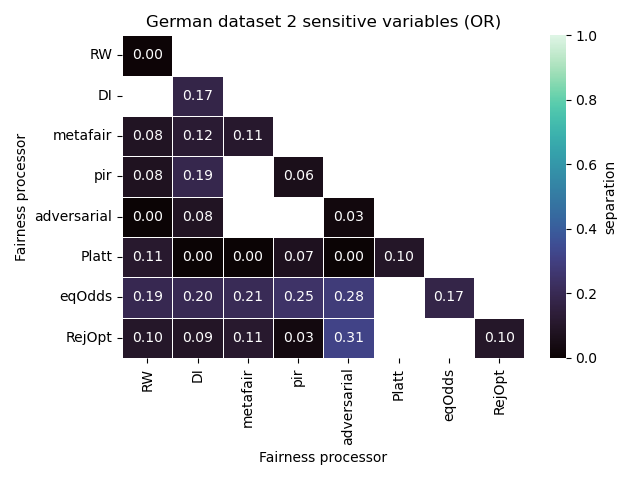}
            \label{fig:matrixGerman2VORsep}
        \end{subfigure}
        \vskip\baselineskip
        \begin{subfigure}[b]{0.49\textwidth}   
            \centering 
            \includegraphics[width=\textwidth]{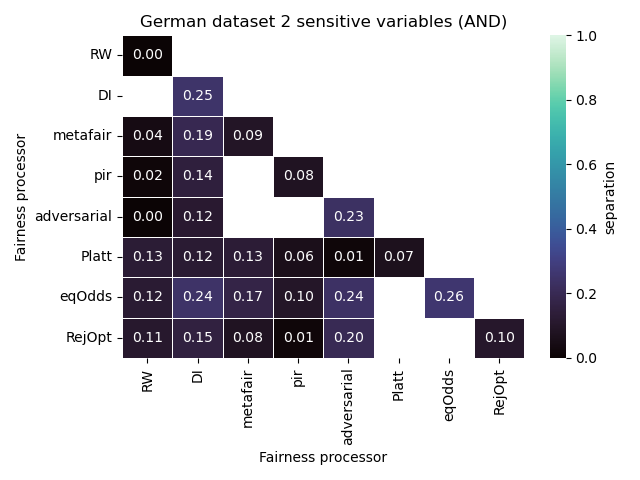}
            \label{fig:matrixGerman2VANDsep}
        \end{subfigure}
        \hfill
        \begin{subfigure}[b]{0.49\textwidth}   
            \centering 
            \includegraphics[width=\textwidth]{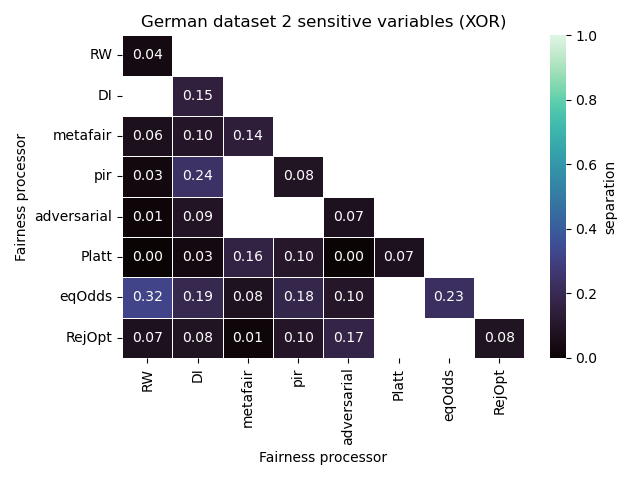}
            \label{fig:matrixGerman2VXORsep}
        \end{subfigure}
        \caption[Separation heatmap for the German data set]
        {\small Separation heatmap for the German data set} 
        \label{fig:SepGerman}
    \end{figure}

\begin{figure}[h!]
    \centering
        \begin{subfigure}[b]{0.49\textwidth}
            \centering
            \includegraphics[width=\textwidth]{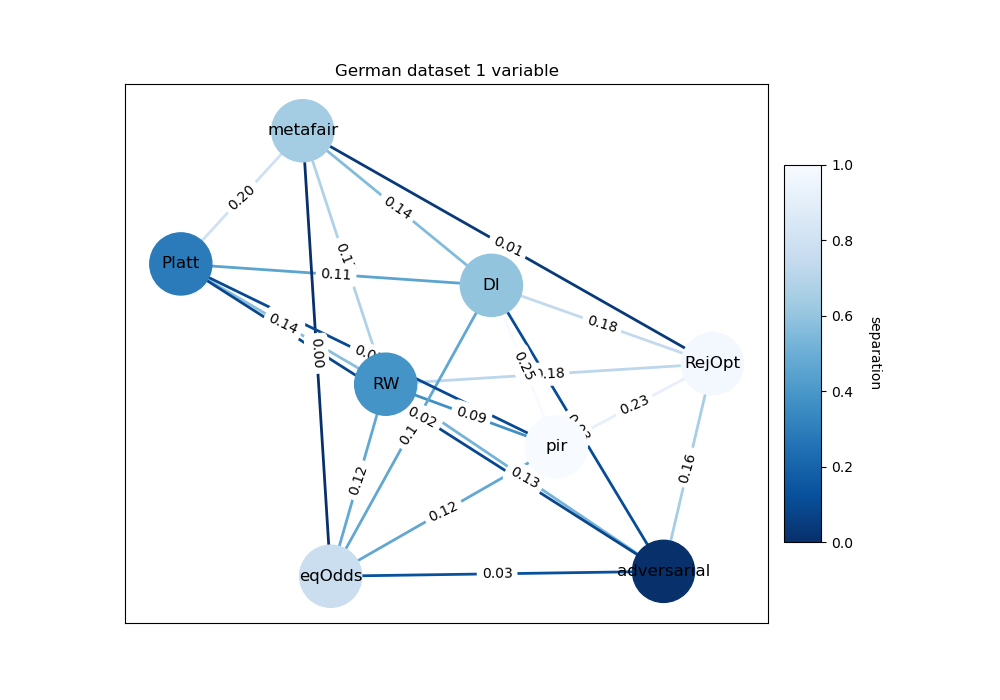}
            \label{fig:graphGerman1Vsep}
        \end{subfigure}
        \hfill
        \begin{subfigure}[b]{0.49\textwidth}  
            \centering 
            \includegraphics[width=\textwidth]{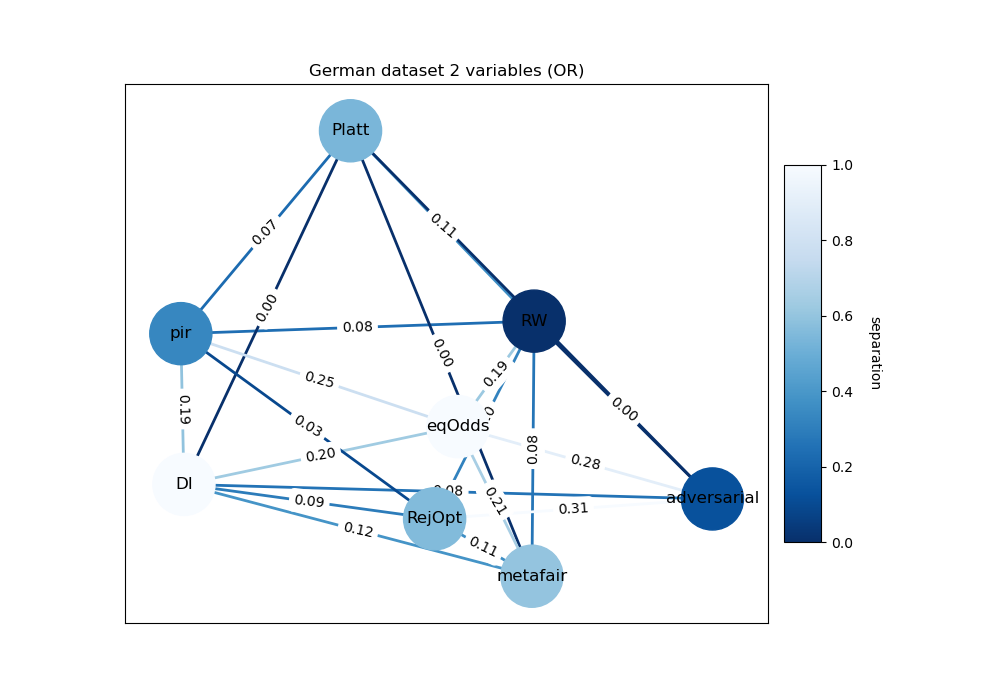}
            \label{fig:graphGerman2VORsep}
        \end{subfigure}
        \vskip\baselineskip
        \begin{subfigure}[b]{0.49\textwidth}   
            \centering 
            \includegraphics[width=\textwidth]{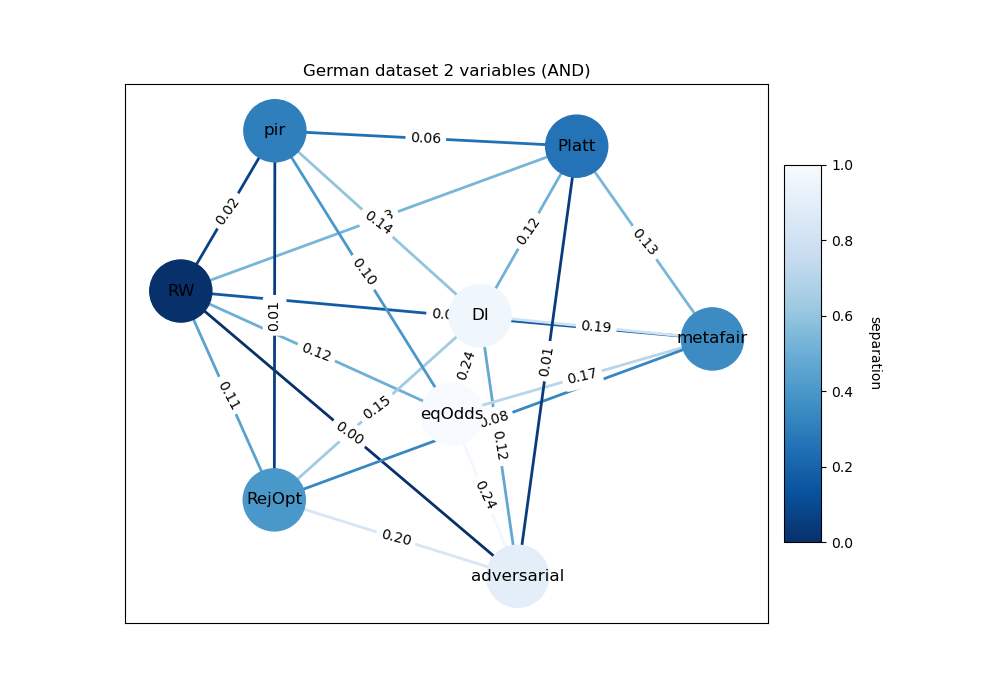}
            \label{fig:graphGerman2VANDsep}
        \end{subfigure}
        \hfill
        \begin{subfigure}[b]{0.49\textwidth}   
            \centering 
            \includegraphics[width=\textwidth]{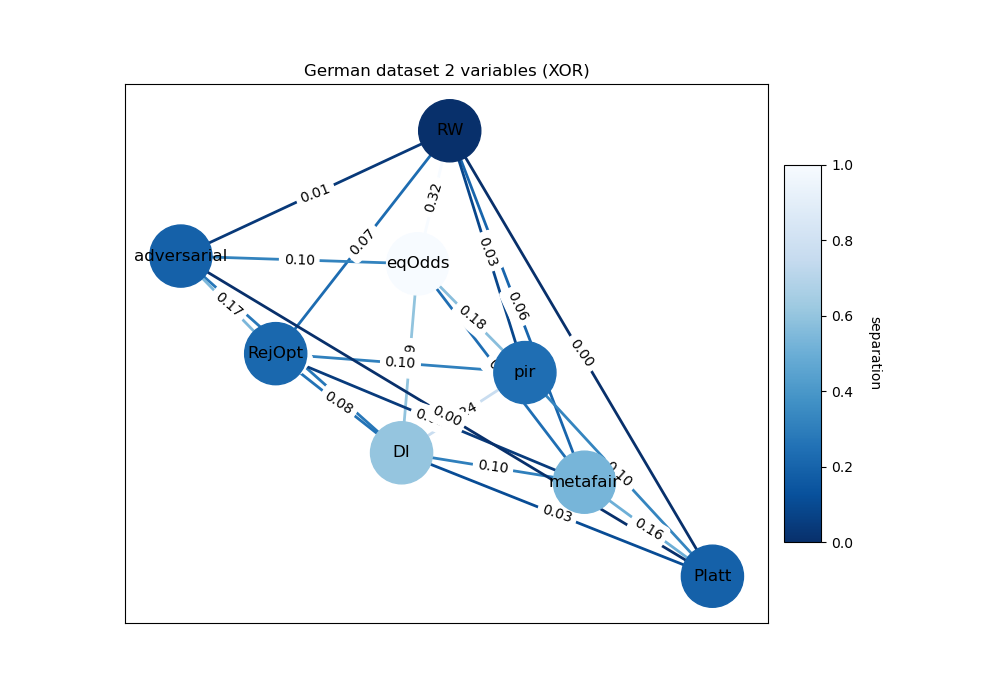}
            \label{fig:graphGerman2VXORsep}
        \end{subfigure}
        \caption[Separation graph in German data set]
        {\small Separation graph in German data set} 
        \label{fig:graphSepGerman}
    \end{figure}

\begin{figure}[h!]
    \centering
        \begin{subfigure}[b]{0.49\textwidth}
            \centering
            \includegraphics[width=\textwidth]{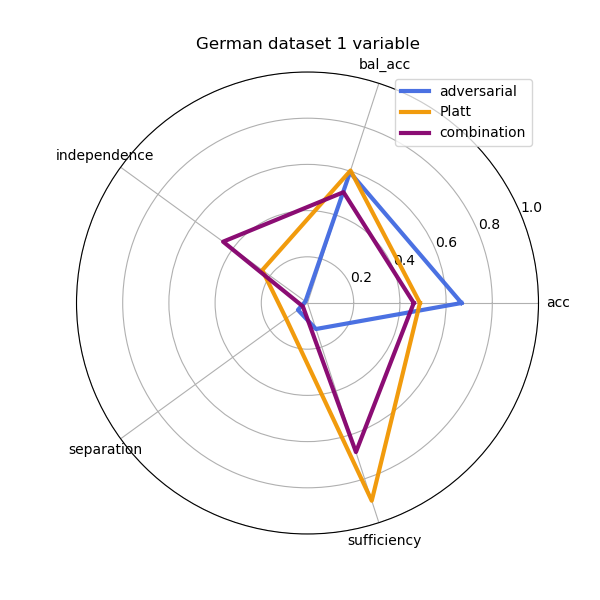}
            \label{fig:radarGerman1V}
        \end{subfigure}
        \hfill
        \begin{subfigure}[b]{0.49\textwidth}  
            \centering 
            \includegraphics[width=\textwidth]{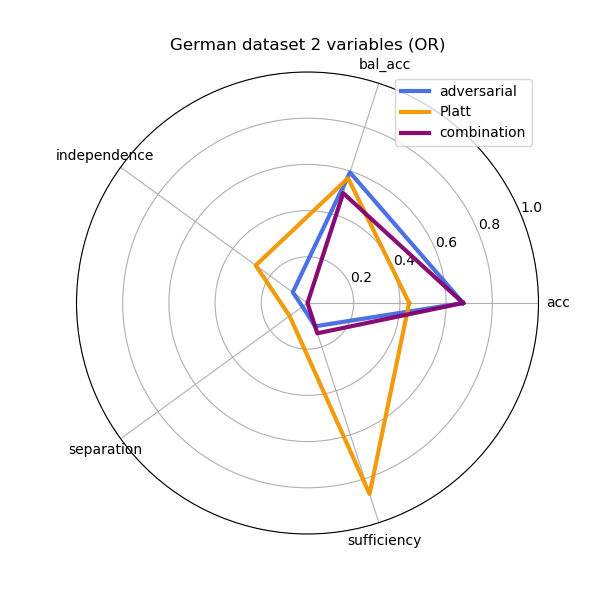}
            \label{fig:radarGerman2VOR}
        \end{subfigure}
        \vskip\baselineskip
        \begin{subfigure}[b]{0.49\textwidth}   
            \centering 
            \includegraphics[width=\textwidth]{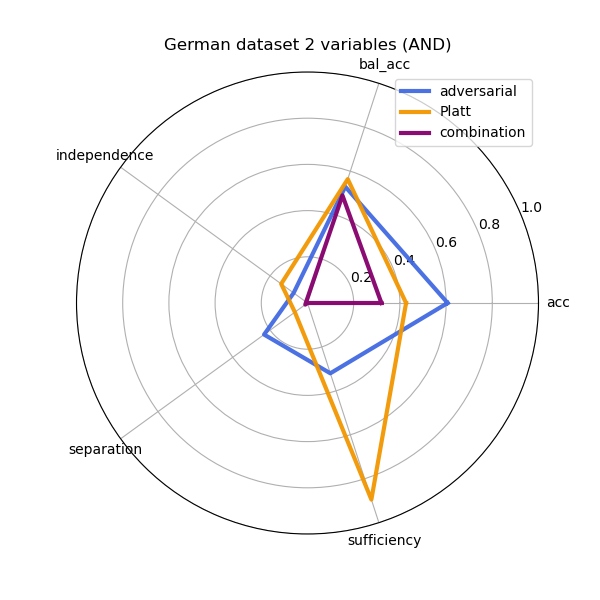}
            \label{fig:radarGerman2VAND}
        \end{subfigure}
        \hfill
        \begin{subfigure}[b]{0.49\textwidth}   
            \centering 
            \includegraphics[width=\textwidth]{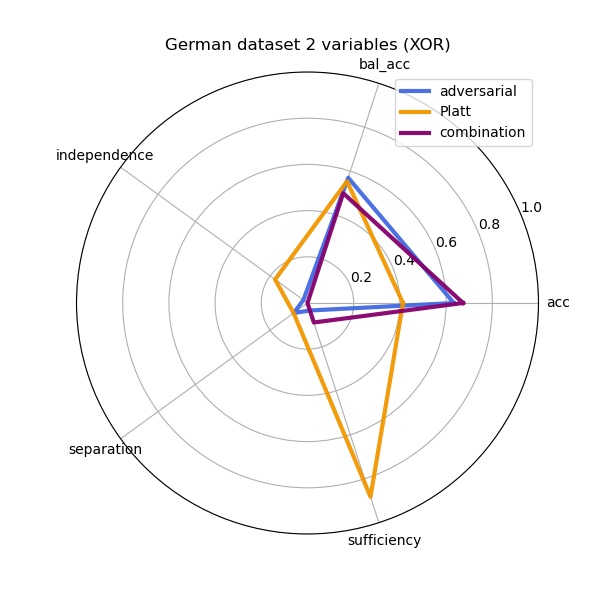}
            \label{fig:radarGerman2VXOR}
        \end{subfigure}
        \caption[Radar plot for the combination of adversarial debiasing and Platt scaling.]
        {\small Radar plot for the combination of adversarial debiasing and Platt scaling.} 
        \label{fig:radarGerman}
    \end{figure}

\begin{figure}[h!]
    \centering
    \includegraphics[width=0.5\linewidth]{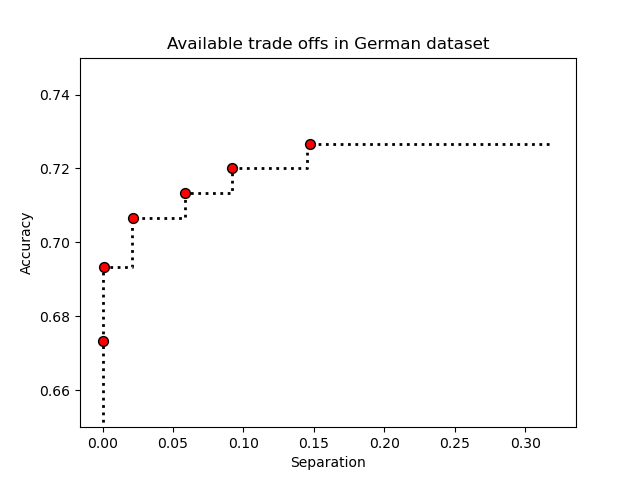}
    \caption{Pareto efficient frontier in German data set.}
    \label{fig:paretoGerman}
\end{figure}

The results are summarized in Figures \ref{fig:AccuracyGerman} to \ref{fig:paretoGerman}. The analysis starts once again by checking the accuracy of the processors, which can be visualized in Figures \ref{fig:AccuracyGerman} and Figure \ref{fig:graphAccGerman}.


In general, the results are much more uniform across the board now: There are no extreme outliers in either direction. Furthermore, it is confirmed that the accuracy of a multistage processor is not bounded above (for instance, the combination, $0.67$, of prejudice index regularizer, $0.61$, and reject option, $0.64$, in the case of one sensitive variable) or below (for instance, the combination, $0.37$, of prejudice index regularizer, $0.61$, and Platt scaling, $0.49$, in the univariate case) by the performance of its constituents. This confirms the previous insights. Furthermore, the most accurate processor is a multistage processor in some scenarios; take, for instance, the case of a single sensitive variable: The highest accuracy, $0.7$, is obtained by combining the metafair classifier, $0.65$, with Platt scaling, $0.49$. 

PI combinations provide the most consistent performance across all combinations: Both PP and IP procedures can degrade accuracy to levels near $0.37$, but the worse performing PI processors are around $0.5$, the worst one being the combination of disparate impact and adversarial debiasing at $0.48$. In the univariate case half of the PI methods improve the performance of its constituents. In the multivariate case, however, the results show more variability, with the choice of LP having a great effect on performance. Both OR and AND processors seem to preserve accuracy, even improving it in some cases (take, for instance, the combination of DI and metafair). On the other hand, the XOR LP produces highly variable results: this method is responsible for the most accurate PI in the combination of disparate impact remover and the prejudice index regularizer and the worst performing in disparate impact remover and adversarial debiasing. PP methods perform best in the univariate scenario. However, at least half of all the processors have an accuracy of $0.5$ or less when using a LP. The worst performing methods in this category are generally those which involve Platt scaling, which seems to degrade performance. Nonetheless, some combinations still show promise, like those involving equal odds, which generally preserves accuracy and it even improves it when coupled with disparate impact in the XOR case. IP processors, on the other hand, showcase synergistic relationships between different fairness methods. Take for instance combinations involving Platt scaling. In general, Platt scaling is the processor with worst perfomance, but when paired with  with the metafair classifier it always improves its accuracy and sometimes it even surpasses the performance of the in-processor. This relationship manages to make this combination the best performer in the individual case, and manages good results in all other scenarios. However, a combination can also prove itself detrimental. Such is the case of the other IP methods involving Platt scaling, which can degrade accuracy to levels around $0.3$. In any case, the results of IP procedures are a mixed bag and depend mostly on the choice of post-processor. This conclusion can be extended into PP combinations and is another instance of the dominating behavior that was observed in the simulation study, although it is now more tamed. In general, combinations involving Platt scaling can show great variability and greatly compromise accuracy in most cases, although some particular combinations may lead to exceptional results. MP involving equal odds lead to an overall good performance, while those involving reject option somewhat compromise fairness in the multivariate case while keeping the accuracy consistent.

The performance of the classifier does not vary as wildly as it did in the simulation study when using a LP. This means that logical processors are an appropriate tool for dealing with multiple sensitive variables because they allow us to reach similar accuracy levels to those of the single variable case. However, the choice of the logical operation can have a great impact on perfomance: Take, for instance, the multistage processor that combines adversarial debiasing with Platt scaling. Using the OR LP improves performance by around $40\%$, while the AND procesor yields a loss of nearly $30\%$. Overall, the results are more consistent when using the OR LP.


The study now proceeds with the analysis of separation. It relies on Figure \ref{fig:SepGerman} and Figure \ref{fig:graphSepGerman}. Once again fairness is not compromised when using MP or LP. Moreover, their use leads to improved fairness in some cases. This is particularly positive in cases where the MP of two average-performing methods produce an exceptionally fair processor; take, for instance, the combination, $0.03$, of the prejudice index regularizer, $0.27$, with Platt scaling, $0.11$, in the case of a single sensitive variable. Furthermore, these MPs can be top performers both in terms of fairness and accuracy; take adversarial debiasing and Platt scaling in the case of two sensitive attributes with OR LP, whose metrics can be seen in Figure \ref{fig:radarGerman}.

In general, MPs discriminate the least in the multivariate case, the correct choice depending on each processor. PI combinations reduce discrimination the most when they involve reweighting. The worst pair of methods is in this cathegory is given by the MP consisting of the disparate impact remover and the prejudice index regularizer, both in the univariate case, $0.25$, and when using XOR, $0.24$. The best couple on the other hand is given by reweighting and adversarial debiasing with both OR, $0.00$, and AND, $0.00$, LPs.

The choice of post-processor seems to have a great impact on the separation of PPs and IPs. The worst combinations arise when using equal odds while the best ones mostly involve Platt scaling, although it is important to keep in mind the accuracy trade-off. The PP combination of reweighting and equal odds processor produces the worst result overall in terms of separation, while the most biased IP results from adversarial debiasing and reject option. In general, results in fairness are more uniform than they were in accuracy.

Now we will analyze the interaction of one single MP, adversarial debiasing and Platt scaling, through the radar graphs found in Figure \ref{fig:radarGerman}. The choice of LP affects the results: the combination of method outperforms both of its constituents in terms of accuracy when using the OR and XOR LPs. However, it underperforms both of them in the other two scenarios. In any case, separation is always improved when using this combination, which suggests that the interaction of these two processors greatly enhances fairness in this metric.

Finally, the available accuracy-separation trade-offs can be visualized through the Pareto efficient frontier in Figure \ref{fig:paretoGerman}, which shows that it is possible to obtain an improvement in separation of nearly $33.3\%$ with a loss of less than $2\%$ in accuracy; in other words, it is feasible to barely compromise performance while obtaining big improvements in fairness.\\


\section{Conclusions and further research}\label{sec:further_research}

This paper has set to introduce two novel techniques in the context of fair ML for credit scoring. 

The first contribution are logical processors or LPs, which provide a way of easily handling multiple sensitive attributes. The results drawn from both the simulation and the case study suggest that this method preserves both accuracy and fairness, which makes them appropriate for the task at hand. Furthermore, they are easy to deploy and can be implemented as a pre-processing step, which facilitates the developement of new fair algorithms without having to worry about whether or not they can be applied to multiple sensitive variables. Nonetheless, this paper only explored the bivariate case, which is why it is important to research the scalability of this tool to handle more than two sensitive variables.

The second contribution is that of multistage processors or MPs, which attack multiple stages of the fair pipeline in order to improve fairness and accuracy. In particular, we have looked for synergistic combinations of methods that have already been established in the literature. The results show that these new processors allow for new trade-offs in fairness and performance, which is essential for decision-makers and provides new ways of dealing with unfair data or models. Furthermore, some instances of these MPs lead to improved separation and accuracy, outperforming the existing methods.

Finally, the findings of this study can be extended to domains well outside of credit scoring: The ideas we introduced are not exclusive to that task. Furthermore, other domains might benefit from the convenience that LPs provide to handle multiple sensitive variables or from the new fairness solutions found in MPs.

Some potential directions for extending our research include:

\subsection{More processors}

This paper considers a limited sample of multistage and logical processors. However, it remains to be seen if the results could be improved by using new, more original multistage processors that exploit the framework in unique ways other than simply stacking fairness processors together. On the other hand, although we have shown the effectiveness of logical processor in handling multiple sensitive variables, it remains to be seen how they generalize when dealing with three or more sensitive attributes. Furthermore, when handling more than two protected variables the possible amount of logical processors dramatically increases, which begs the question of how to make an appropriate choice.

\subsection{Interpretation and explanations}

The results shown are positive and allow us to form certain intuitions about multistage processors. However, many questions remain about the relationships and interactions between fairness procedures: does applying this particular pre-processor improve the predictions of this in-processor? Why or why not? In what scenarios? Answers to these questions could further our understanding of why and how MPs work. On the other hand, the conclusions drawn for LPs seem optimistic, but perhaps when dealing with other sensitive attributes, such as a race or religion, the results may change for sociological reasons. A more thorough review of the consequences of logical processors in more diverse contexts could enlighten our understanding of logical processors.

\subsection{Hyper-parameter tuning}

The paper has studied the interaction of different fairness processors with fixed hyper-parameters. However, this means that one direction left unexplored is to check whether or not these new multistage processors can outperform their individual counterparts if their hyper-parameters are tuned. Furthermore, the results would show what the optimal configurations are when combining processors, which should shed some light into precisely how they interact.

\section*{Acknowledgments}
This research is part of the I + D + i projects PDC2022-133359, PID2022-137243OB-I00 and  PID2022-137818OB-I00 funded by Ministerio
de Ciencia, Innovación y Universidades/AEI/10.13039/501100011033 and European Union NextGenerationEU/PRTR. This initiative has also been partially carried
out within the framework of the Recovery, Transformation and Resilience Plan funds, financed by the European Union (Next Generation) through the grant ANTICIPA
and the ENIA 2022 Chairs for the creation of university-industry chairs in AI-AImpulsa: UC3M-Universia.

\appendix
\section{Analyzing an individual simulation instance}

\begin{figure}[h!]
        \centering
        \begin{subfigure}[b]{0.49\textwidth}
            \centering
            \includegraphics[width=\textwidth]{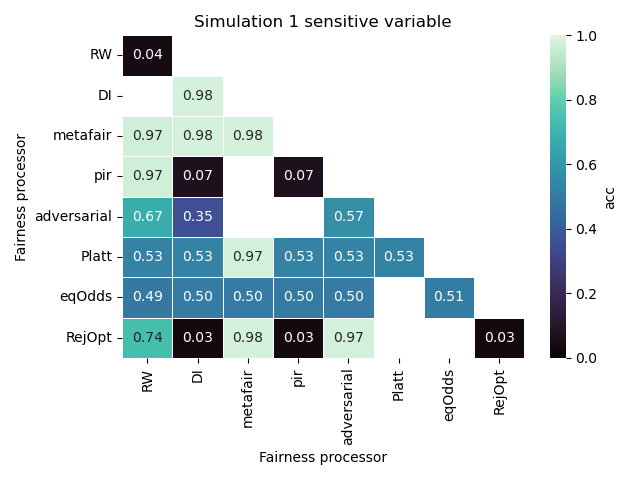}
            \label{fig:matrixSimul1Vacc}
        \end{subfigure}
        \hfill
        \begin{subfigure}[b]{0.49\textwidth}  
            \centering 
            \includegraphics[width=\textwidth]{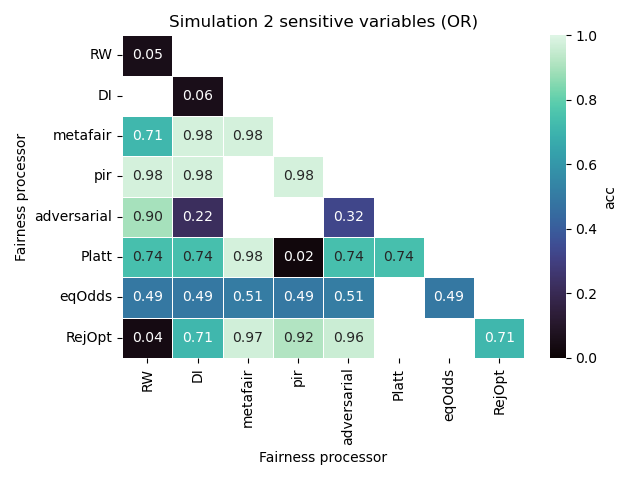}
            \label{fig:matrixSimul2VORacc}
        \end{subfigure}
        
        \begin{subfigure}[b]{0.49\textwidth}   
            \centering 
            \includegraphics[width=\textwidth]{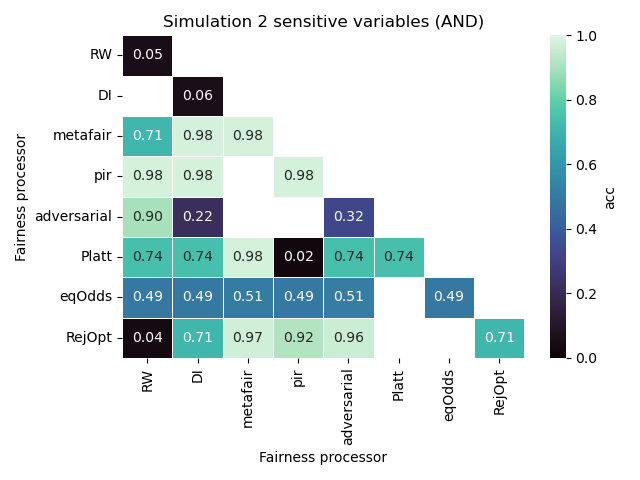}
            \label{fig:matrixSimul2VANDacc}
        \end{subfigure}
        \hfill
        \begin{subfigure}[b]{0.49\textwidth}   
            \centering 
            \includegraphics[width=\textwidth]{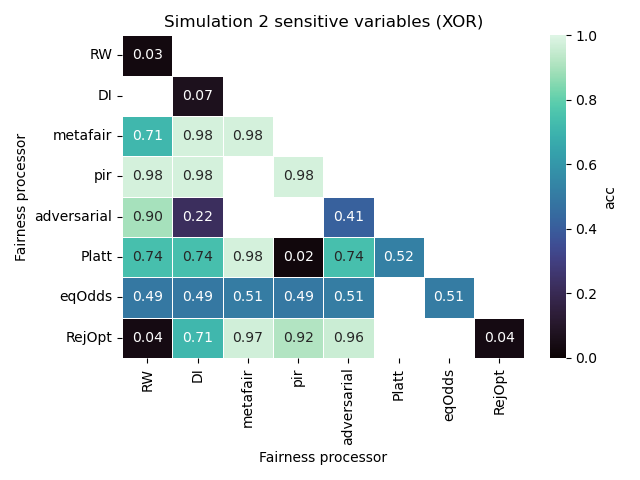}
            \label{fig:matrixSimul2VXORacc}
        \end{subfigure}
        \caption[Accuracy of multistage processors in the simulation study]
        {\small Accuracy of multistage processors in a single instance of the simulation study.} 
        \label{fig:AccuracySimul}
    \end{figure}

\begin{figure}[h!]
    \centering
        \begin{subfigure}[b]{0.49\textwidth}
            \centering
            \includegraphics[width=\textwidth]{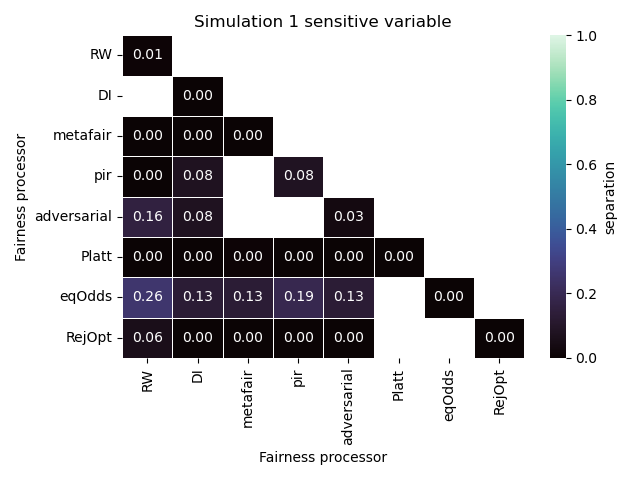}
            \label{fig:matrixSimul1Vsep}
        \end{subfigure}
        \hfill
        \begin{subfigure}[b]{0.49\textwidth}  
            \centering 
            \includegraphics[width=\textwidth]{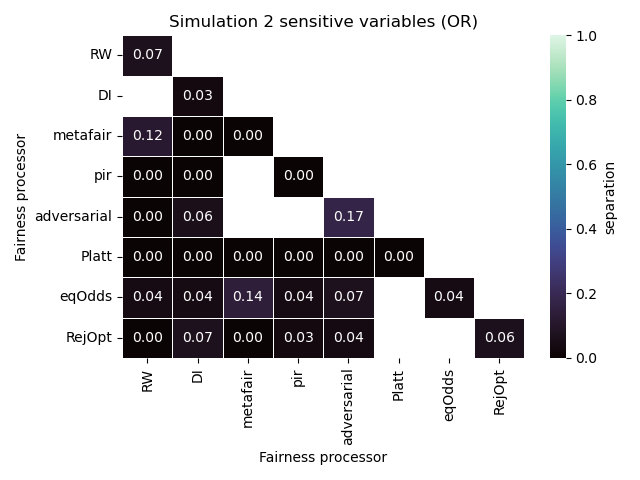}
            \label{fig:matrixSimul2VORsep}
        \end{subfigure}
        \vskip\baselineskip
        \begin{subfigure}[b]{0.49\textwidth}   
            \centering 
            \includegraphics[width=\textwidth]{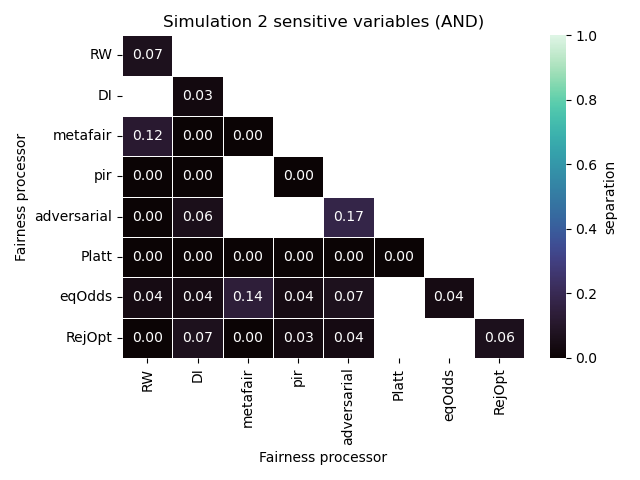}
            \label{fig:matrixSimul2VANDsep}
        \end{subfigure}
        \hfill
        \begin{subfigure}[b]{0.49\textwidth}   
            \centering 
            \includegraphics[width=\textwidth]{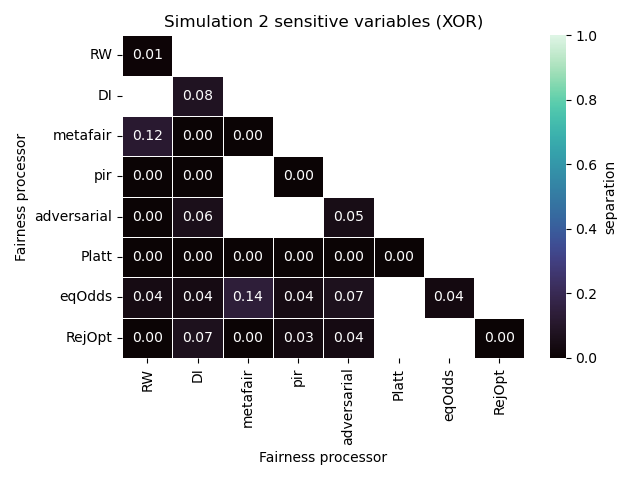}
            \label{fig:matrixSimul2VXORsep}
        \end{subfigure}
        \caption[Separation heatmap for the instance of the simulation study.]
        {\small Separation heatmap for the instance of the simulation study.} 
        \label{fig:SepSimul}
    \end{figure}

\begin{figure}[h!]
    \centering
    \includegraphics[width=0.5\linewidth]{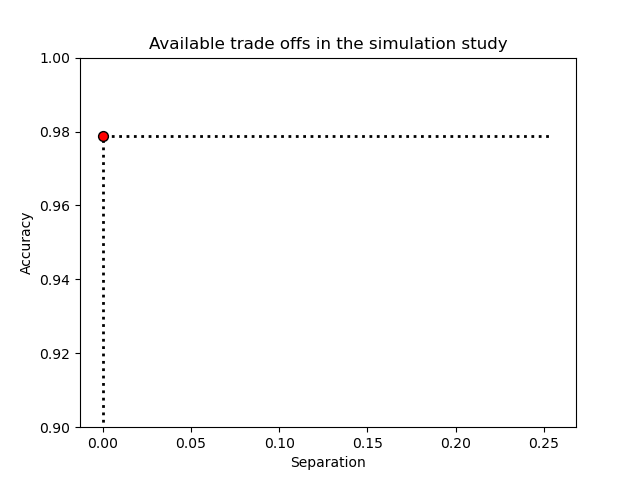}
    \caption{Pareto efficient frontier in the simulation study.}
    \label{fig:paretoSimul}
\end{figure}

We include the results for a single instance of the simulation study in order to isolate the performance and fairness of the methods studied in this settings, with the idea of better grasping the behavior of the processors on a single data set. These results are summarized in Figure \ref{fig:AccuracySimul} to Figure \ref{fig:paretoSimul}. We start the analysis by checking the accuracy of the methods, which can be visualized in Figure \ref{fig:AccuracySimul}. 

At first glance, the simulation seems to be a resounding success: the combined use of fairness processors can lead to great improvements in performance. This is the case of the MP consisting of reweighting and prejudice index regularizer. These two processors are some of the worst performers in the univariate case with an accuracy level of less than $0.1$. However, the MP that combines them reaches one of the best accuracies of the batch, reaching $0.97$. However, there are some MPs that underperform with respect to their constituents, although they seem to be a minority.

The use of a post-processor (both PP and IP) seems to have a great impact on the performance of the resulting MP, exhibiting the dominating behavior we previously speculated. This is specially clear in the case of the equal odds processor, where every MP involving it seems to have an accuracy of around $0.5$.

The choice of logical processor does not seem to make a big impact on the overall quality of the predictions of the classifiers, but it is a great cause of variation for certain processors. For instance, Platt scaling seems to work best when using the AND or OR procedures, while adversarial debiasing performs best when using XOR processor. 


The separation of the methods can be visualized in Figure \ref{fig:SepSimul}. In general, a great level of fairness is achieved in all situations: the level of separation is generally close to $0$.
In any case, the use of MPs can also lead to improved fairness. In particular, the combination of reweighting and the prejudice index regularizer in the univariate case seems particularly promising. Table \ref{tab:poc_2} shows how their interaction leads to one of the best solutions, being a top performer both in terms of accuracy (it presents the second best accuracy of the univariate case) and fairness (achieving perfect separation) even though their constituents show poor accuracy.

\begin{table}[h!]
    \centering
    \begin{tabular}{c | c c c}
        Processor & Accuracy & Fairness  \\
        \hline
        Reweighting & $0.04$ & $0.01$ \\
        Prejudice index regularizer & $0.07$ & $0.08$ \\
        Combination & $0.97$ &  $0.00$
    \end{tabular}
    \caption{Performance and fairness metrics for the combination of reweighting and prejudice index regularizer in the univariate case.}
    \label{tab:poc_2}
\end{table}

Turning our attention to LPs, their use does not compromise fairness, and it even improves separation for the equal odds processors. In any case, the choice of LP does not seem to matter in this case, being responsible only for small differences in certain processors like adversarial debiasing.
    
Finally, we can visualize the available fairness-accuracy trade-offs through the Pareto efficient frontier in Figure \ref{fig:paretoSimul}. Indeed, the flat shape of the frontier indicates the lack of trade-offs available, which is a consequence of the high levels of fairness and accuracy that can be achieved. 


 \bibliographystyle{elsarticle-num}
 \bibliography{cas-refs}





\end{document}